\documentclass[journal,twocolumn,transmag,review = true]{IEEEtran}
\usepackage{caption}
\usepackage{times}
\usepackage{epsfig}
\usepackage{graphicx}
\usepackage{amsmath}
\usepackage{amssymb}
\usepackage{multirow}
\usepackage{subfigure}
\usepackage{ulem}
\usepackage{color}
\usepackage{enumitem}
\usepackage{epstopdf}
\usepackage{float}

\usepackage[backref]{hyperref}
\usepackage{makecell}

\ifCLASSINFOpdf
\else
\fi
\begin{document}
%
\title{TBNet:Two-Stream Boundary-aware Network for Generic Image Manipulation Localization}

\author{Zan Gao, IEEE Member, Chao Sun, Zhiyong Cheng, IEEE Member, \\ Weili Guan, Anan Liu, Senior Member, IEEE and Meng Wang, IEEE Fellow}

\markboth{Journal of Latex}%
{Shell \MakeLowercase{\textit{et al.}}: Bare Demo of IEEEtran.cls for IEEE Transactions on Magnetics Journals}
%





\maketitle
\begin{abstract}
Abstract - Finding tampered regions in images is a hot research topic in machine learning and computer vision. Although many image manipulation location algorithms have been proposed, most of them only focus on the RGB images with different color spaces, and the frequency information that contains the potential tampering clues is often ignored. Moreover, among the manipulation operations, splicing and copy-move are two frequently used methods, but as their characteristics are quite different,  specific methods have been individually designed for detecting the operations of either splicing or copy-move, and it is very difficult to widely apply these methods in practice. To solve these issues, in this work, a novel end-to-end two-stream boundary-aware network (abbreviated as TBNet) is proposed for generic image manipulation localization in which the RGB stream, the frequency stream, and the boundary artifact location are explored in a unified framework. Specifically, we first design an adaptive frequency selection module (AFS) to adaptively select the appropriate frequency to mine inconsistent statistics and eliminate the interference of redundant statistics. Then, an adaptive cross-attention fusion module (ACF) is proposed to adaptively fuse the RGB feature and the frequency feature. Finally, the boundary artifact location network (BAL) is designed to locate the boundary artifacts for which the parameters are jointly updated by the outputs of the ACF, and its results are further fed into the decoder. Thus, the parameters of the RGB stream, the frequency stream, and the boundary artifact location network are jointly optimized, and their latent complementary relationships are fully mined. The results of extensive experiments performed on four public benchmarks of the image manipulation localization task, namely, CASIA1.0, COVER, Carvalho, and In-The-Wild, demonstrate that the proposed TBNet can significantly outperform state-of-the-art generic image manipulation localization methods in terms of both MCC and F1 while maintaining robustness with respect to various attacks. Compared with DeepLabV3+ on the CASIA1.0, COVER, Carvalho, and In-The-Wild datasets, the improvements in MCC/F1 reach 11\%/11.1\%, 8.2\%/10.3\%, 10.2\%/11.6\%, and 8.9\%/6.2\%, respectively\footnote{Manuscript received Jan 29th, 2021; This work was supported in part by the National Natural Science Foundation of China (No.61872270, No.62020106004, No.92048301, No.61572357).  Young creative team in universities of Shandong Province (No.2020KJN012), Jinan 20 projects in universities (2020GXRC040). New Artificial Intelligence project towards the integration of education and industry in Qilu University of Technology (No.2020KJC-JC01). Tianjin New Generation Artificial Intelligence Major Program (No.18ZXZNGX00150, No.19ZXZNGX00110).

Z. Gao, C. Sun (Corresponding Author) and Z.Y Cheng (Corresponding Author) are with Shandong Artificial Intelligence Institute, Qilu University of Technology (Shandong Academy of Sciences), Jinan, 250014, P.R China. Z. Gao is also with Key Laboratory of Computer Vision and System, Ministry of Education, Tianjin University of Technology, Tianjin, 300384, China.

Weili Guan is with the Faculty of Information Technology, Monash University Clayton Campus, Australia.

A.A. Liu is with the School of Electrical and Information Engineering, Tianjin University, Tianjin 300072, China. 
 
M. Wang is with the school of Computer Science and Information Engineering, Hefei University of Technology, Hefei, 230009, P.R China. 
}.

\end{abstract}

\begin{figure}[ht]
\begin{center}
\includegraphics[width=3.5in,height = 2.0in]{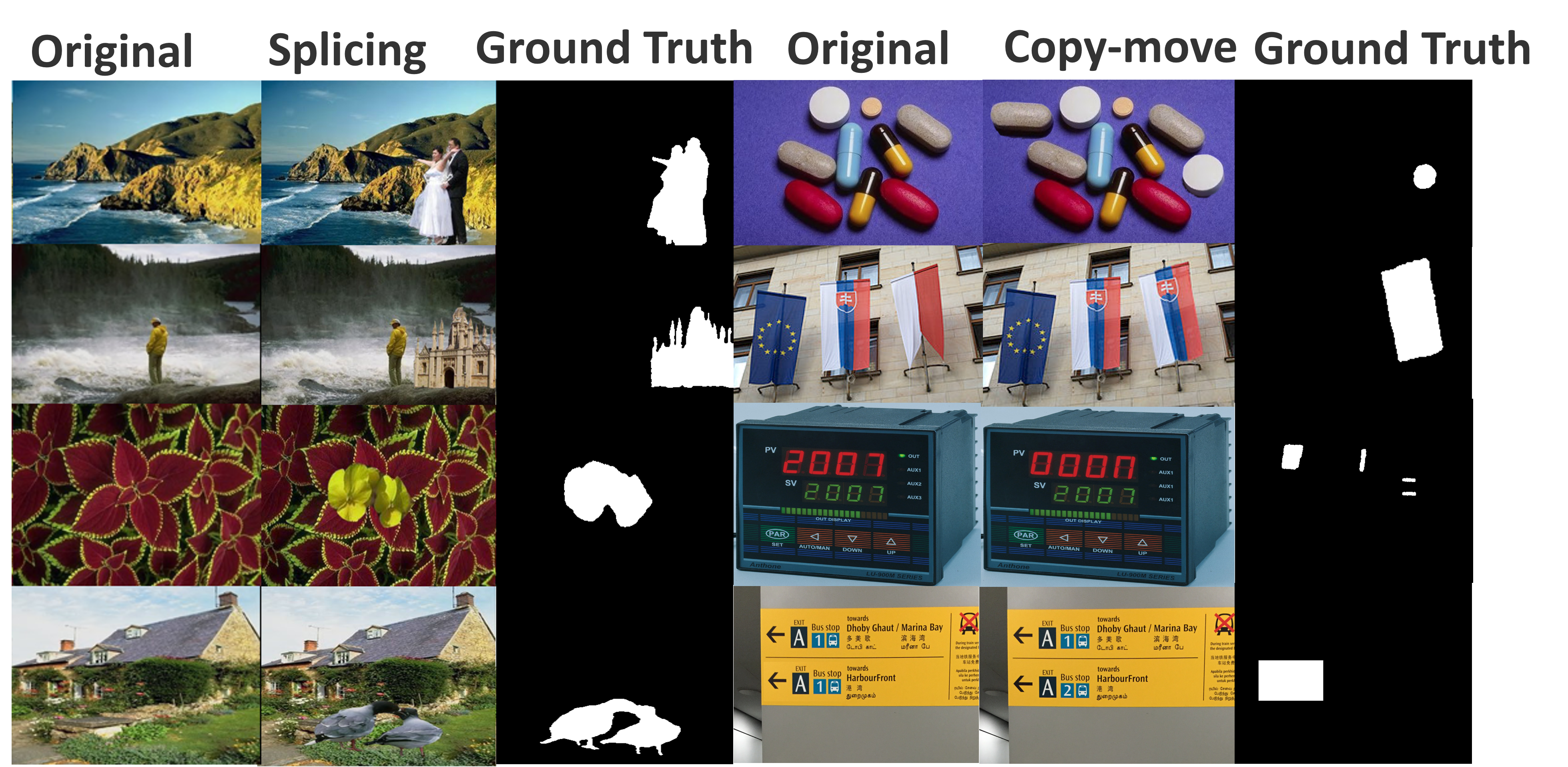}
\caption{Examples of manipulated images and ground truth masks. Among the manipulation operations, splicing and copy-move are two frequently used methods. Images from the top-left three columns belong to the splicing, and images from the top right three columns belong to the copy-move. }
\label{fig:1}
\end{center}
\vspace{-1.5em}
\end{figure}

\begin{IEEEkeywords}
Generic Image Manipulation Localization; Two-Stream Boundary-aware; Adaptive Frequency Selection; Adaptive Cross-attention Fusion; Boundary Artifact Localization; 
\end{IEEEkeywords}

\maketitle

\IEEEdisplaynontitleabstractindextext

%
\IEEEpeerreviewmaketitle

\section{Introduction}
%
%
%
%

\IEEEPARstart
The rapid development and use of multimedia collection equipment have ushered in an era of multimedia information explosion. Almost everyone can produce and transmit a large number of digital images. Moreover, the simple operation of image editing software has made image manipulation increasingly easy, enabling ordinary people to easily process and modify image content. However, image tampering changes the original expression of the image and delivers misleading and false information. Fig. 1 shows several examples of manipulated images and ground truth masks. It is observed from Fig. 1 that the tampered image looks natural, and it is difficult to find the clues indicating that the images were tampered with. Since the dissemination of this information may damage national and personal security, finding the tampered region is a highly important task in image forensics.

To date, many researchers have proposed different approaches for the detection of image tampering \cite{Bunk2017DetectionAL,Park2018DoubleJD,Bappy2019Hybrid, Bappy2017ExploitingSS,Mohammed2018BoostingIF,Zhou2020GenerateSA}, but these methods have focused on the RGB or other color spaces. For example, double JPEG compression features, re-sampling features, or other hand-crafted physical features \cite{Bunk2017DetectionAL,Park2018DoubleJD,Bappy2019Hybrid} are extracted from RGB images, but with the development of tampering technology, hand-crafted features can easily be shielded in a targeted manner. Therefore, some researchers \cite{Bappy2017ExploitingSS,Mohammed2018BoostingIF,Zhou2020GenerateSA} have employed convolutional neural networks (CNN) to automatically learn tampering features from the RGB or other color spaces. Although these deep learning features often can obtain better performance than that of the hand-crafted features, they often fail when encountering images with complex tampering manipulations. This is because they only focus on the RGB or other color spaces, but the frequency domain that may contain potential clues indicating tampering in the frequency statistics is often ignored.

In a different approach, some researchers \cite{Zhou2018LearningRF, Bappy2019Hybrid, Fridrich2012RichMF, Cozzolino2015SplicebusterAN, Cozzolino2017RecastingRL,Rao2016ADL,Liu2018ImageFL, Zhou2018LearningRF} have demonstrated the utility of the frequency information, and have proposed different methods to exploit such information; for example, Zhou et al. \cite{Zhou2018LearningRF} focused on the combination of the RGB domain and the high-frequency noise domain, with the latter containing the information loss of the tampering clues; Fridrich et al. \cite{Fridrich2012RichMF} proposed a rich steganalysis model (SRM) to obtain the local noise features that endowed the image with high-frequency information sensitivity. SRM is quite helpful for distinguishing the tampered regions from the real regions \cite{Cozzolino2015SplicebusterAN}, and many researchers \cite{Rao2016ADL,Liu2018ImageFL, Zhou2018LearningRF} have employed SRM to detect the tampering image. In these approaches, we observe that the noise domain is a high-frequency sensitive image that has been preprocessed by a simplified SRM high-pass filter to remove the 
-frequency information; thus, some low-frequency information is directly abandoned and may contain some clues to help identify the tampered region. Moreover, the latent complementary relationships between the RGB features and the frequency features also need to be further mined.

Moreover, among the manipulation operations \cite{Wu2018ImageCF, Wu2018BusterNetDC, Islam2020DOAGANDA}, splicing and copy-move are two frequently used methods, but their characteristics are quite different; for example, the manipulation operation of the splicing copies a region from an image and then pastes it to another image (the second column in Fig.1), while the manipulation operation of the copy-move is to copy and paste an image area within the same image (the fifth column in Fig. 1). Moreover, post-processing using methods such as gfiltering, illumination, and blurring, is often performed to conceal the traces of tampering. Based on the use of these methods, we can observe that the true and false regions of a spliced image have different statistical characteristics of different images, however, the copy-move is an internal tampering operation of the same image. Therefore, the statistical features of the real area and the tampered area will be very similar in this case, and the features designed to identify the splicing area cannot be used here. If the refined post-processing is performed, it will be more difficult to detect the tampered regions; thus, some unique approaches are individually designed for detecting the operation of splicing or copy-move \cite{Liu2018DeepFN, Bi2019RRUNetTR, Xiao2020ImageSF, Liu2020ExposingSF, Wu2018ImageCF, Zhong2020AnED, Wu2018BusterNetDC, Barni2021CopyMS, Islam2020DOAGANDA}; however, these approaches are difficult to apply widely in practical use.

To solve these issues, in this work, a novel end-to-end TBNet network is proposed for generic image manipulation localization; in TBNet, the RGB features, frequency features, and boundary artifacts location are discussed in a unified framework. To fully take advantage of the frequency, the AFM module is designed to adaptively select the appropriate frequency. Then, to mine the latent complementary relationships between the RGB feature and the frequency feature, the ACF model is developed. Finally, to locate the boundary artifacts, BAL is employed for which the parameters are jointly optimized with the RGB and frequency streams.

In summary, the main contributions of this paper are fourfold:
\begin{itemize} 

\item We develop a novel end-to-end TBNet module for the generic image manipulation localization task in which the RGB stream, the frequency stream, and the boundary artifacts location are explored in a unified framework. In this module, the tampering manipulations of splicing and copy-move can be simultaneously detected.

\item We design an AFS module to adaptively select the appropriate frequency to mine inconsistent statistics and eliminate the interference of redundant statistics, and then an ACF module is proposed to adaptively fuse the RGB feature and the frequency feature.

\item We propose a BAL network to locate the boundary artifacts for which the parameters are jointly updated by the outputs of the ACF, and its results are further fed into the Decoder. Thus, the parameters of the RGB stream, the frequency stream, and the boundary artifacts location network are jointly optimized, and their latent complementary relationships are fully mined.


\end{itemize}

The remainder of the paper is organized as follows. Section II introduces the related work, and Section III describes the proposed TBNet method. Section IV describes the experimental settings and the analysis of the results. Section V presents the details of the ablation study, and the concluding remarks are presented in Section VI.

\section{Related Work}
Image manipulation localization is a hot research topic, and in particular splicing and copy-move are two frequently used operations for image manipulation. Since the characteristics of different manipulation operations are quite different, several unique approaches have been individually designed for detecting the operation of splicing or copy-move; these approaches are called Restricted Manipulation Localization \cite{Liu2018DeepFN, Bi2019RRUNetTR, Xiao2020ImageSF, Liu2020ExposingSF, Wu2018ImageCF, Zhong2020AnED, Wu2018BusterNetDC, Barni2021CopyMS, Islam2020DOAGANDA}. However, since these approaches are difficult to use in a wide range of practical applications, some researchers have proposed the generic manipulation localization methods \cite{Bappy2017ExploitingSS, salloum2018image, Chen2018EncoderDecoderWA, Zhou2020GenerateSA}. Thus, in this section, these two aspects are described separately.

\subsection{Restricted Manipulation Localization }

Among the manipulation operations, splicing and copy-move are two frequently used methods. The tampered images obtained by these two manipulation operations look quite natural and identical to real images particularly when the post-processing such as gfiltering, illumination, and blurring, is performed to cover up the traces of tampering; however, their characteristics are quite different, motivating the design of several specific approaches for detecting the operations of splicing \cite{Liu2018DeepFN, Bi2019RRUNetTR, Xiao2020ImageSF, Liu2020ExposingSF} or copy-move \cite{Wu2018ImageCF, Zhong2020AnED, Wu2018BusterNetDC, Barni2021CopyMS, Islam2020DOAGANDA}. For example, Wu et al. \cite{Wu2018ImageCF} proposed an end-to-end image copy-move tampering detection framework where the correlation between each pair of pixels in the feature map is calculated, and the similar regions are selected according to the correlation size; additionally, the high-level features of the similar regions are extracted and fed into the Decoder; Wu et al. \cite{Wu2018BusterNetDC} further expanded the framework and integrated the strengths of the boundary artifact and the regional similarity to detect the tampered regions; Zhong et al. \cite{Zhong2020AnED} proposed a Dense-InceptionNet module to autonomously learn the feature correlations and search the possible forged snippets through the matching clues; Barni et al. \cite{Barni2021CopyMS} designed a multi-branch CNN architecture to solve the source-target disambiguation problem by revealing the presence of interpolation artifacts and boundary inconsistencies in the copy-moved area. Islam et al. \cite{Islam2020DOAGANDA} proposed a dual-order attention module to focus on similar regions and generate a more accurate mask using the generative adversarial network (GAN).

Similarly, for the detection of the splicing manipulation, Liu et al. \cite{Liu2018DeepFN} proposed a new deep fusion neural network (fusion-net) to locate the tampered area by tracking the network boundary, and then a small number of images are used to fine-tune the network so that the fusion network can distinguish whether or not an image block was synthesized from different sources; Bi et al. \cite{Bi2019RRUNetTR} proposed an end-to-end RRU-Net module to achieve the splicing detection without any preprocessing and post-processing; the core idea of this approach was to improve the learning of CNN by the propagation and the feedback process of the residual in CNN. Zhou et al. \cite{Zhou2020GenerateSA} proposed a splicing detection model based on the structure of GAN. Xiao et al. \cite{Xiao2020ImageSF} proposed a C2RNet module that cascaded the coarse convolutional neural network and the fine convolutional neural network and then extracted the image attribute difference between the untampered area and the tampered area from the image patches of different scales, but only the high-level visual features are extracted while low-level features rich in forensic clues were ignored in this method. To solve this problem, Liu et al. \cite{Liu2020ExposingSF} proposed a new neural network to focus on learning low-level forensic features and detect the splicing forgery; the high-level forensic features are ignored in their method. However, in practice, when facing unknown images in practical applications, the type of image tampering cannot be known in advance preventing the wide practical application of these restricted manipulation localization approaches.

\subsection{Generic Manipulation Localization}
To solve the above issues, the generic manipulation localization algorithms have been proposed; for example, Liu et al. \cite{Liu2018DeepFN} used different types of basic networks to improve the manipulation localization performance, where the networks complement each other; Zhou et al. \cite{Zhou2018LearningRF} proposed an RGB-N framework that transformed the passive forensics task of image content into the target detection task; Bappy et al. \cite{Bappy2019Hybrid} proposed an encoder-decoder architecture based on resampling features to solve the pixel-level positioning problem of the tampered image regions; Bappy et al. \cite{Bappy2017ExploitingSS} designed a module to capture the anomalous characteristics of the boundary between the modified regions and the non-modified regions; Salloum et al. \cite{salloum2018image} proposed a multi-task image passive forensics framework (MFCN) based on edge enhancement that predicted the edge mask and the segmentation mask, and then the intersection of the boundary mask after the hole filling and the segmentation mask was taken as the tampered area. Chen et al. \cite{Chen2018EncoderDecoderWA} proposed the Deeplabv3+ module where the encode-decode structure was utilized and had achieved good results on pixel-level semantic segmentation tasks; Zhou et al. \cite{Zhou2020GenerateSA} proposed a GSR-Net module based on the semantic segmentation, that consisted of Generate, Segment and Replace. However, in most of these approaches, the frequency information \cite{Bappy2017ExploitingSS, salloum2018image, Chen2018EncoderDecoderWA, Zhou2020GenerateSA} is often ignored, or only the high-frequency information \cite{Bappy2019Hybrid, Zhou2018LearningRF} is explored for detecting the manipulation. Moreover, the boundary artifacts location is also quite helpful for detecting the tampered regions, so that some researchers \cite{salloum2018image,Zhou2020GenerateSA} also have explored this approach, but the complementary relationship between the boundary artifacts location and the tampered regions needs to be further mined.

\section{TBNet: Two-Stream Boundary-aware Network for Generic Image Manipulation Localization}
Since the frequency information is often ignored or the low-frequency information is directly removed in most existing image manipulation localization methods, but the frequency information may contain some key clues to help identify the tampered regions, in this work, a novel end-to-end two-stream network architecture including the RGB information and the frequency information is first built; then, when an image is tampered by the splicing or copy-move, some subtle traces on the edge of the manipulation are often left; such boundary artifacts are very useful, and therefore in this work, the boundary artifacts network is further embedded into the two-stream network. Most importantly, the parameters of the RGB stream, the frequency stream, and the boundary artifacts location network are jointly optimized, and their latent complementary relationships are fully mined. Thus, a novel end-to-end TBNet module is proposed for generic image manipulation localization. The network architecture of the TBNet is shown in Fig. 2 that consists of the RGB stream, the frequency stream, and the boundary artifacts location network. In these streams, ResNet101 with well-known basic network architecture is used as the backbone, and therefore, in the following, we will introduce the adaptive frequency selection module and the adaptive cross-attention fusion module, respectively. Moreover, we also present the boundary artifacts location network. Finally, the loss function of the TBNet is given. All of the characteristics are described below.
\begin{figure*}[htb]
\begin{center}
\includegraphics[width=7in,height =3.5in]{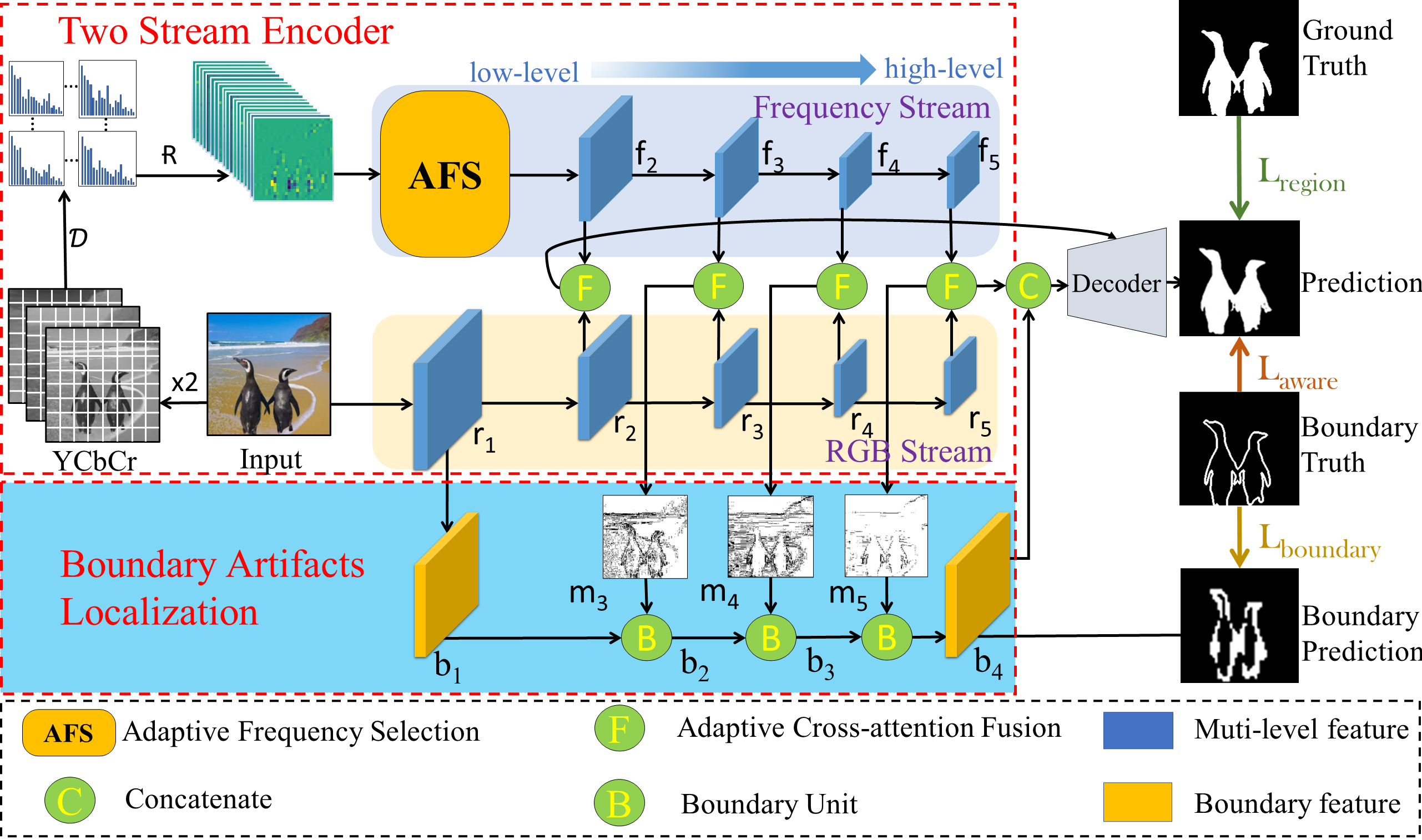}
\caption{The network architecture of the TBNet. It mainly consists of the two-stream encoder (the RGB stream and the frequency stream are included), and the boundary artifacts location network. In the frequency stream, an image with the RGB color space is first converted to the YCbCr color space, and then the image is divided into many patches. $\mathcal{D}$ is the discrete cosine transform that is used to transform these patches into the frequency domain. $\mathcal{R}$ is the procedure of the reshape and normalization. `AFS' is an adaptive frequency selection module that adaptively selects the appropriate frequency channels after adjusting the shape, and `F' indicates an adaptive cross-attention fusion module to mine the latent complementary relationships between the RGB features and the frequency features, and fuse them. In TBNet, ResNet101 is utilized as the backbone, and $f_2$, $f_3$, $f_4$ and $f_5$ are the layers of the conv2-x, conv3-x, conv4-x and conv5-x in ResNet101, respectively; similarly, $r_1$, $r_2$, $r_3$, $r_4$ and $r_5$ are the layers of the conv1-x, conv2-x, conv3-x, conv4-x and conv5-x in ResNet101, respectively. $b_1$, $b_2$, $b_3$ and $b_4$ are the inputs and outputs of the corresponding boundary units. We note that if necessary, bilinear interpolation can be used to upsample or downsample the Image or feature maps. The parts highlighted by the red dashed line are our key contributions (best viewed in color).
}
\end{center}
\vspace{-1.5em}
\end{figure*}
\begin{figure}[htb]
\begin{center}
\includegraphics[width=3.5in,height = 2.0in]{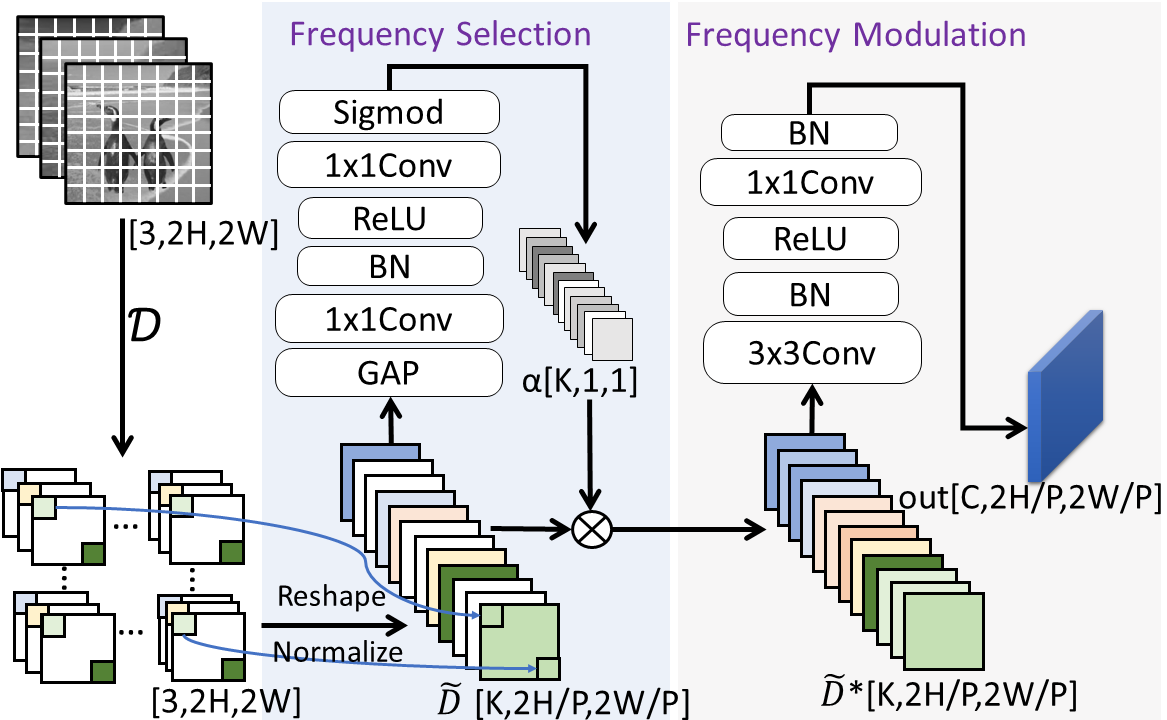}
\caption{Pipeline of the AFS module that consists of a frequency selection layer (FS) and a frequency modulation layer (FM).}
\label{fig:3}
\end{center}
\vspace{-2.5em}
\end{figure}

\subsection{Adaptive Frequency Selection (AFS) Module}
Since frequency information may contain key image tampering clues, some methods have used it \cite{Zhou2018LearningRF, Bappy2019Hybrid, Fridrich2012RichMF, Cozzolino2015SplicebusterAN, Cozzolino2017RecastingRL,Rao2016ADL,Liu2018ImageFL}; however, in these approaches, the high-frequency information is often used, and the low-frequency information is simply ignored. Thus, in this work, we develop an adaptive frequency selection (AFS) module to adaptively select the appropriate frequency information to mine inconsistent statistics and eliminate the interference of redundant statistics. Our method includes two components, namely, a frequency selection layer and a frequency modulation layer. The former is used to adaptively assign the weight to each frequency, and the latter is utilized to exchange the information between the different frequency channels and adjust the channel dimensions to embed the backbone. The pipeline of the AFS module is given in Fig. 3. In the following, we separately introduce the frequency selection layer and the frequency modulation layer.

Suppose that there exists an RGB image $I\in \mathbb{R}^{3\times H\times W}$; then, we first convert and resize it to the YCbCr image $I' \in \mathbb{R}^{3\times 2H\times 2W}$ (The height $H$ and the width $W$ of the image $I$ are resized to $2H$ and $2W$ in order to keep the same size as that of the RGB stream in the subsequent frequency processing), and then $P\times P$ blocks of three components of Y, Cb, and Cr are built. Moreover, we further perform Discrete cosine transform ($\mathcal{D}$) \cite{Ahmed1974Discrete} for them to obtain $P \times P$ frequency coefficients. As shown in Fig. 3, each coefficient represents different frequencies. For each component, the coefficients of the same frequency are placed in the same channel according to the position relationship and expressed as $d_k\in \mathbb{R}^{2H/P\times 2W/P}$, and the dct coefficient in $d_k$ retains relative position relationship. $\widetilde{d}_k$ is the normalized version of each frequency $d_k$ obtained by subtracting its mean $\mu_k$ and dividing by its standard deviation $\sigma_k$ as shown in Eq \ref{eq:1}. It is defined as
\begin{equation}
\begin{aligned}
& \widetilde{d}_k=(d_k-\mu_k)/\sigma_k, \label{eq:1}
\end{aligned}
\end{equation}
Thus, we mark the reshaped and normalized data as $\widetilde{D}=\{\widetilde{d}_1,\widetilde{d}_2,...,\widetilde{d}_K\} \in \mathbb{R}^{K\times 2H/P\times 2W/P}$. In our experiments, $P=8$, $K=192$. Since different channels of the input features are at different frequencies, we conjecture that some frequency channels have less manipulation information, so that performance should be improved by decreasing the weights of these trivial frequency channels. Inspired by squeeze-and-excitation block (SE-Block) \cite{Hu2018SqueezeandExcitationN} that utilizes the channel-wise information to emphasize the informative features and suppress the trivial features, we propose a learning-based frequency selection mechanism to adaptively recalibrate frequency responses by explicitly modeling interdependencies between different frequency channels. Thus, we can exploit the relative importance of each input frequency channels, and then we employ it to assign a score $\alpha_k\in[0,1]$ to each frequency channel. $\alpha$ can be computed as:
\begin{equation}
\begin{aligned}
& \alpha= \sigma\{Conv_2(\delta(\beta(Conv_1(g(\widetilde{D})))))\},
\end{aligned}
\end{equation}
where $g(\widetilde{D})=\frac{1}{2H/P\times 2W/P}\sum_{i=1}^{2H/P}\sum_{j=1}^{2W/P}\widetilde{D}_{[:,i,j]}$ is the global average pooling (GAP). $\delta$ denotes the Rectified Linear Unit (ReLU), and $\beta$ denotes the Batch Normalization (BN). $\sigma$ is the Sigmoid function. The kernel sizes of $Conv_1$ are $(K/r)$$ \times K\times1\times1$ and $Conv_2$ is $K\times (K/r)$ $\times1 \times1$ where $r$ is a dimensionality reduction hyperparameter to reduce the amount of calculation. In our experiment $r=8$. Let $\alpha=\{a_1,a_2,...,a_K\}$ is the weights of the input frequency channels $\widetilde{D}=\{\widetilde{d}_1,\widetilde{d}_2,...,\widetilde{d}_K\}$. The learning-based frequency selection can be defined as follows:
\begin{equation}
\begin{aligned}
& \widetilde{d}^*_k=\widetilde{d}_k\odot a_k,k\in[1,K],
\end{aligned}
\end{equation}
where $\odot$ is the element-wise product, and $d^*$ is the weighted frequency channel. The output of the frequency selection layer can be represented by $\widetilde{D}^*$=$\{\widetilde{d}^*_1,\widetilde{d}^*_2,...\widetilde{d}^*_K\}\in \mathbb{R}^{K\times 2H/P\times 2W/P}$.

Since the frequency selection layer must be embedded into the backbone, but its dimension of the frequency channel is different, in our experiments, we design a frequency modulation layer to replace the conv1-x of the ResNet101 (the backbone); this layer can be computed as
\begin{equation}
\begin{aligned}
O_{AFS}=FM(\widetilde{D}^*)=\beta\{Conv_4(\delta(\beta(Conv_3(\widetilde{D}^*))))\},
\end{aligned}
\end{equation}
where $Conv_3$ indicates the $3 \times 3$ group convolution and group = $K$. The kernel sizes of $Conv_4$ are $C\times K \times 1 \times1$, and $O_{AFS} \in\mathbb{R}^{C\times 2H/P\times 2W/P}$ is the output of the AFS module. In our experiments, $P$ is set to $8$; thus, $O_{AFS} \in\mathbb{R}^{C\times H/4\times W/4}$ can be easily embedded into the backbone.

\subsection{Adaptive Cross-attention Fusion Module (ACF)}
In the TBNet, the RGB feature and the frequency feature are separately extracted, and these two features are complementary and may show a synergistic performance promotion effect for each other if they can be effectively fused. Therefore in this work, we design an adaptive cross-attention fusion module (ACF) to adaptively fuse them where the cross-attention mechanism is employed, and then the combined feature is used to locate the manipulation region and detect the boundary artifacts. To make full use of their complementarities, the cross-attention mechanism is used to control the information flow of each feature map where these two feature maps are first concatenated, and then these fused feature is utilized to calculate the cross-attention weights, and they can be calculated by
\begin{equation}
\begin{aligned}
G^{r_i}=C_{1\times1}(f_i||r_i),G^{r_i}\in \mathbb{R}^{1\times H_i\times W_i}, \\
G^{f_i}=C^{'}_{1\times1}(f_i||r_i),G^{f_i}\in \mathbb{R}^{1\times H_i\times W_i},
\end{aligned}
\end{equation}
where $i$ is the index of the convolutional layer in the ResNet101. $f_i$ and $r_i$ indicate the $i^{th}$ RGB feature maps and $i^{th}$ frequency feature map, respectively, and $H_i$ and $W_i$ are the height and width of the corresponding feature maps, respectively. $||$ is a concatenation operation, and the cross-feature map can be obtained by the $f_i||r_i$. $C_{1\times1}$ and $C^{'}_{1\times1}$ indicate a $1 \times 1$ convolution operation to the cross-feature map so that the attention weight can be obtained. $G^{r_i}$ indicates the spatial gate (the cross-attention weight) for the RGB feature map, $G^{f_i}$ denotes the spatial gate (the cross-attention weight) for the frequency feature map. We note that two different $1 \times 1$ convolution layers are used for the RGB feature and the frequency feature the parameters of which are individually learned; thus, the values of $G^{r_i}$ and $G^{f_i}$ are different.

To further normalize these cross-attention weights, a softmax function is applied on these two gates, and they can be defined as
\begin{equation}
\begin{aligned}
A^{r_i(h,w)}=\frac{e^{G^{r_i(h,w)}}} {e^{G^{r_i(h,w)}}+e^{G^{f_i(h,w)}}}, \\
A^{f_i(h,w)}=\frac{e^{G^{f_i(h,w)}}} {e^{G^{r_i(h,w)}}+e^{G^{f_i(h,w)}}},
\end{aligned}
\end{equation}
where $h$ is the $h^{th}$ row pixel index, and $w$ is the $w^{th}$ column pixel index. $G^{r_i(h,w)}$ is the weight of the $h^{th}$ row and $w^{th}$ column in the RGB feature. Similarly, $G^{f_i(h,w)}$ denotes the weight of the $h^{th}$ row and $w^{th}$ column in the frequency feature. $A^{r_i(h,w)}$, $A^{f_i(h,w)}$ $\in \mathbb{R}^{1\times H_i\times W_i}$ are the cross-attention weights for the RGB feature and the frequency feature, respectively, and $A^{r_i(h,w)} + A^{f_i(h,w)} =1$. Thus, the fusion feature can be obtained by the following weighted scheme
\begin{equation}
\begin{aligned}
m_i(h,w) = r_i(h,w)\cdot A^{r_i(h,w)}+f_i(h,w)\cdot A^{f_i(h,w)},
\end{aligned}
\end{equation}
where $m_i \in \mathbb{R}^{1\times H_i\times W_i}$ is the $i^{th}$ fused feature map, and $m_i(h,w)$ indicates the value of the $h^{th}$ row and $w^{th}$ column in the $m_i$. Finally, the fused feature is fed into the BAL module that is used to detect the boundary artifacts. Moreover, to maintain their specificities during the long-term transmission, the cross-complementarily aggregate features at a certain position in space are utilized layer-by-layer.

\subsection{Boundary Artifact Localization (BAL)}

The RGB information and the frequency information are quite useful for image tampering region detection, but the tampering techniques are very strong, and the tampered image looks natural so that it is still very difficult for the RGB and frequency information to locate the tampered region. However, the unnaturalness of the edge may provide a strong hint about the existence of tampering, and therefore, boundary artifacts localization is introduced as a related task to finer detection manipulation edge. To date, some researchers \cite{salloum2018image,Zhou2020GenerateSA} have proposed some related boundary artifacts localization approaches, but these approaches are different from the normal edge detection methods \cite{zhou2021Hierarchical, Huang2020Semantic, Han2018Robust}, and the low-resolution image is often used ($r_2$ or $r_3$ or $r_4$ in Fig. 2 is often utilized to detect the boundary artifacts), but some key tampering clues may be lost in the low-resolution image. Moreover, in these approaches, the boundary artifacts localization network
\begin{figure}[htb]
\begin{center}
\includegraphics[width=3.2in,height = 1.9in]{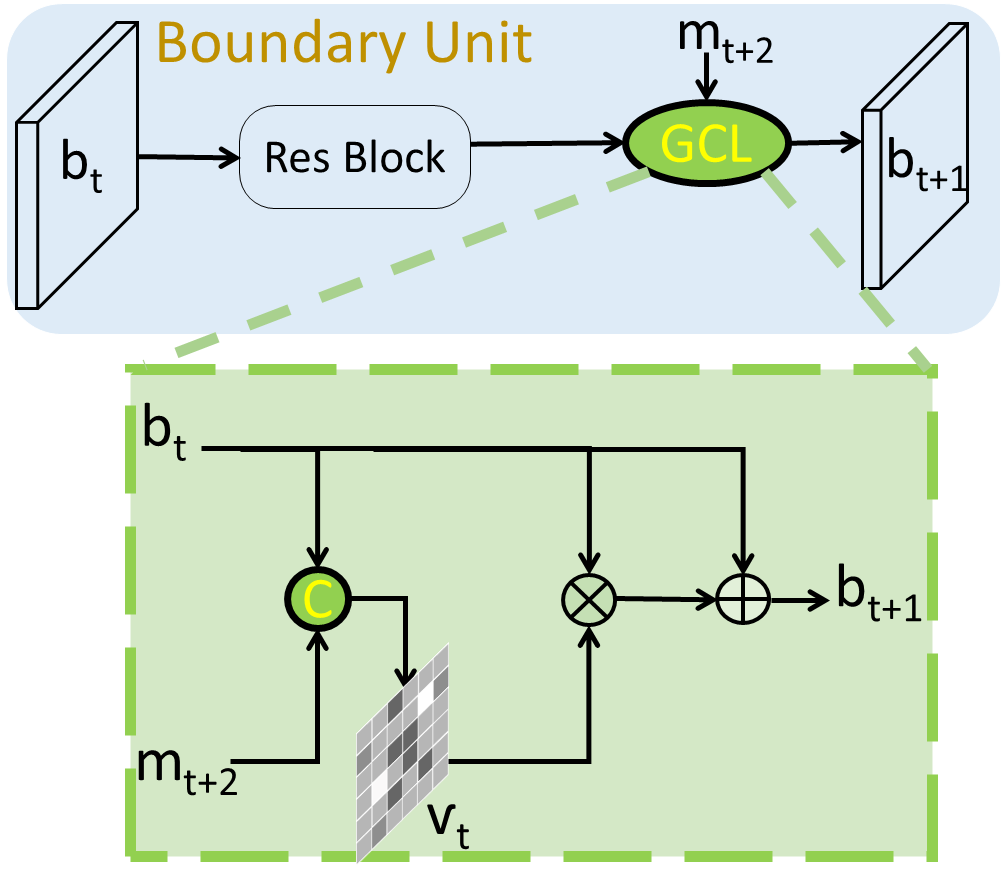}
\caption{Pipeline of the boundary unit. `Res Block' indicates the residual basic block, and ‘GCL’ denotes the gated convolution layer. }
\end{center}
\vspace{-1.5em}
\end{figure}
is often built separately, and the RGB information or the frequency information is often ignored.

Thus, in this work, we propose a BAL module to locate the boundary artifacts in the high-resolution image whose parameters are jointly updated by the outputs of the ACF, and its results are further fed into the Decoder. In this way, the parameters of the RGB stream, frequency stream, and the boundary artifacts location network are jointly optimized, and their latent complementary relationships are well-mined. Its whole network architecture is shown at the bottom of Fig. 2, and the pipeline of the boundary unit is given in Fig. 4. Specifically, since some tampering clues are often lost in the low-resolution feature, the high-resolution image is always used in BAL. In our experiments, the high- resolution image $r_1 \in R^{H/2 \times W/2}$ is directly transferred to $b_1 \in R^{H \times W}$, and $b_1$ is further fed to the residual basic block. Moreover, since the RGB information and the frequency information are also helpful for detecting the boundary artifacts, the fusion feature of the RGB information $r_3$ and the frequency information $f_3$ (the outputs of the ACF) are also employed, but the bilinear interpolation is utilized for upsampling, and the higher level fusion feature $m_3 \in R^{H \times W}$ can be obtained as the high-level prior knowledge that enables the boundary localization to adopt an effective shallow architecture. To fully mine their relationships, these two high-resolution images are further fed into the gated convolution layer (GCL), and the GCL is used to deactivate the activations that are not deemed to be relevant by the higher-level information.

Suppose T is the total number of layers in the BAL module, and $t\in\{1,..., T\}$ is the layer index of the BAL. $m_{t+2}$ and $b_t$ indicate the fusion feature and the output of the residual basicblock of the $t^{th}$ layer in the BAL module, respectively. In the GCL, $m_{t+2}$ and $b_t$ are first connected together, and then a normalized $1 \times 1$ convolutional layer $C_{1\times1}$ and a sigmoid function $\sigma$ are employed, so that we can obtain an attention map $\upsilon_t\in\mathbb{R}^{H\times W}$. Intuitively, $\upsilon_t$ is regarded essentially as an attention map that gives more weights to these areas with important boundary information, and $\upsilon_t$ can be defined
\begin{equation}
\begin{aligned}
& \upsilon_t = \sigma\{C_{1\times1}(m_{t+2}||b_t)\},
\end{aligned}
\end{equation}
After obtaining the attention map $\upsilon_t$, an element-wise product between $\upsilon_t$ and $b_t$ is applied, and then a residual connection and a channel-wise kernel weight $w_t$ are followed, and the outputs $b_{t+1}$ of the GCL can be defined
\begin{equation}
\begin{aligned}
& b_{t+1}^{(i,j)} = \{(b_t^{(i,j)}\upsilon_t^{(i,j)})+b_t^{(i,j)}\}^Tw_t,
\end{aligned}
\end{equation}

Then, $b_{t+1}$ is fed into the next boundary unit, and the fusion feature is also employed. In our experiments, $T$ is set to $4$. To further mine the relationships between the boundary artifacts and the tampering regions, the outputs of the BAL module is further fed into the Decoder. Thus, the parameters of the RGB stream, the frequency stream, and the boundary artifacts location network are jointly optimized, and their latent complementary relationships are well-mined. Moreover, we note that when we optimize the BAL parameters, the ground-truth binary boundary artifacts mask obtained by subtracting the erosion of the dilated binary ground truth mask is used to supervise the optimization. We will introduce this in the next section.

\subsection{Loss Function}
Since the tampered region detection and the boundary artifacts detection can be mutually reinforcing, the parameters of RGB stream, frequency stream and the boundary artifacts location network are jointly optimized, and their latent complementary relationships can be well mined. Thus, the loss function can be defined as
\begin{equation}
\begin{aligned}
& L=\lambda_1 L_{region}+\lambda_2 L_{boundary}+\lambda_3 L_{aware}
\end{aligned}
\end{equation}
where $\lambda_1$, $\lambda_2$ and $\lambda_3$ are the trade-off parameters to balance the contributions of each term. $L_{region}$ and $L_{boundary}$ indicate the loss of the tampered region prediction and the loss of the boundary prediction, respectively. Since the ratio of the tampered boundary pixels and other pixels is not balanced, the weighted binary cross-entropy loss \cite{Yu2017CASENet} is employed in the $L_{region}$ and $L_{boundary}$. To further accurately locate the critical boundray artifact pixels, $L_{aware}$ is designed that computes the loss between the predicted label and the ground truth only at the edge pixels, and the boundary-aware loss $L_{aware}$ is computed by:
\begin{equation}
\begin{aligned}
L_{aware}= -\frac{1}{HW}\sum_{i=1}^{HW}[u_i=1]\{w_1\cdot\widehat{p_i}\cdot log(p_i)\\+w_2\cdot(1-\widehat{p_i})\cdot log(1-p_i)\},
\end{aligned}
\end{equation}
where $i$ is the index and $u_i$ denotes the edge label. $\widehat{p_i}$ indicates the ground truth label of the region map, and $p_i$ denotes the predicted label of the region map. $[*]$ denotes when $u_i$ is equal to one, its results is set to either 1 or 0.

\section{Experiments and Discussions}
To assess the performance of the TBNet, we performed experiments on four public image manipulation localization datasets, namely, CASIA \cite{Dong2013CASIAIT}, Carvalho \cite{Khan2015ExposingDI}, COVER \cite{Wen2016COVERAGEA}, and In-The-Wild \cite{Huh2018FightingFN}. The remainder of this section is organized as follows: 1) four public generic image manipulation localization datasets are introduced, 2) the competitors in the experiments are described, 3) the experimental settings are described, and 4) the performance evaluation and comparison on these four public datasets are described.

\subsection{Datasets}
\begin{itemize}
\item CASIA \cite{Dong2013CASIAIT}: CASIA has two versions. CASIA1.0 contains
921 manipulated images including splicing and copy-move. The objects are carefully selected to match the context in the background. Cropped regions are subjected to post-processing including rotation, distortion, and scaling. CASIA2.0 is a more complicated dataset with 5123 images where Manipulations include splicing and copy-move. Postprocessing techniques such as filtering and blurring are applied to make the regions visually realistic, and the manipulated regions cover animals, textures, and natural scenes. In our experiments, CASIA2.0 and part of the synthesized dataset \cite{Bappy2019Hybrid} are used to train the TBNet and then test it on the CASIA1.0 dataset.

\item Carvalho \cite{Khan2015ExposingDI}: Carvalho is a manipulation dataset designed to conceal illumination differences between manipulated regions and authentic regions. The dataset contains 100 splicing images and all of the manipulated objects are people. Contrast and illumination are adjusted through post-processing.

\item COVER \cite{Wen2016COVERAGEA}: COVER focuses on copy-move manipulation and has 100 images. The manipulation objects are used to cover similar objects in the original authentic images and thus are challenging for humans to recognize visually without close inspection.

\item In-The-Wild \cite{Huh2018FightingFN}: It is a splicing dataset containing 201 online splicing images. The manipulation region is usually people, objects, and animals. The tampered region can be well integrated into the background after careful cutting. The manipulated image has a strong visual impact, and the size of the tampered region is inconsistent.
\end{itemize}
\vspace{-0.8em}

\subsection{Competitors}
In our experiments, the following popular algorithms were compared:
\begin{itemize}
\item DeepLabV3+ (Baseline) \cite{Chen2018EncoderDecoderWA}: Since our basic network architecture is inspired by the DeepLabV3+, it is used as our baseline model that is directly used to detect the image manipulation, and its backbone is also ResNet101.

\item NoI \cite{Mahdian2009UsingNI}: NOI models the local noise through wavelet coefficients and predicts that the local noise in manipulated regions is inconsistent with the authentic area.

\item CFA \cite{Ferrara2012ImageFL}: CFA estimates the camera internal CFA mode of each patch in the image, and segments the CFA characteristic abnormal area.

\item RGB-N \cite{Zhou2018LearningRF}: RGB-N transforms the tamper location into an object detection task, and then a two-stream Faster R-CNN for which the inputs are the RGB stream and noise stream is built to make the final prediction.

\item EXIF-consistency \cite{Huh2018FightingFN}: A self-consistent approach is proposed for manipulation localization where the metadata is used to learn useful feature.

\item MFCN \cite{salloum2018image}: MFCN predicts the edge mask and the segmentation mask, and the intersection of the boundary mask and the segmentation mask after the hole filling is taken as the tampered area.

\item GSR-Net \cite{Zhou2020GenerateSA}: This approach is an image manipulation based on the semantic segmentation that consists of Generate, Segment, and Replace. It has obtained the best performance on many public image manipulation datasets \cite{Dong2013CASIAIT, Khan2015ExposingDI, Wen2016COVERAGEA}.
\end{itemize}

\vspace{-0.8em}
\subsection{Implementation Details}
Since the training samples are very limited, we first randomly overlap-slip the holistic image in the CASIA2.0 dataset, to obtain a set of image blocks with dimensions of $256 \times 256$. To obtain much more training data, these blocks are randomly flipped. In this way, 46,000 image blocks can be obtained. Moreover, 54,000 image blocks (27,000 images for the copy-move manipulation and 27,000 image blocks for the splicing manipulation) are also randomly selected from the synthetic dataset \cite{Bappy2019Hybrid}, and these images are also resized to $256 \times 256$; thus, in total, 100,000 image blocks are used to train the networks. In our experiments, ResNet101 is used as the backbone of the TBNet, and it is pre-trained by the ImageNet dataset with a learning rate of 0.001, and then TBNet is trained by these 100,000 training samples with a learning rate of 0.01. For the optimization, the SGD optimizer is employed to optimize the objective function where the momentum is set to 0.9 and the weight-decay is set to $5e^{-4}$. Moreover, $\lambda_1$, $\lambda_2$, and $\lambda_3$ are empirically set to 0.05, 0.05, and 0.9, respectively for all of the datasets. For the other models, the default modules (codes) or the experimental results are used. Moreover, since our basic network architecture is inspired by the DeepLabV3+, for a fair comparison, ResNet101 is also used as its backbone, and therefore, the same training strategy is used as that for TBNet. Finally, since the F1 and Matthews Correlation Coefficient (MCC) metrics are often utilized as the evaluation metrics in the image manipulation localization algorithms \cite{Zhou2020GenerateSA, Zhou2018LearningRF}, we also strictly follow these metrics in our experiments.
\begin{table*}
\centering
\fontsize{10}{10}\selectfont
\caption{Performance evaluation and comparison on four benchmarks. ‘-’ denotes that the result is not available in the literature, and the values shown in black indicate the best performance in each column.}
\renewcommand{\arraystretch}{1.5}
\begin{tabular}{ccccccccc}
\hline
\multirow{1}{*}{Datasets} & \multicolumn{2}{c} {CASIA1.0}& \multicolumn{2}{c} {COVER}& \multicolumn{2}{c} {Carvalho}& \multicolumn{2}{c} {In-The-Wild} \\
\cline{1-9}
Metrics &MCC &F1 &MCC &F1 &MCC &F1 &MCC &F1\\
\hline
NOI \cite{Mahdian2009UsingNI} &0.180 &0.263 & 0.107 & 0.269 &0.255 &0.343 &0.159 &0.278 \\
CFA \cite{Ferrara2012ImageFL} &0.108 &0.207 &0.050 &0.190 &0.164 &0.292 &0.144 &0.270 \\
MFCN \cite{salloum2018image} &0.520 &0.541 &- &- &0.408 &0.480 &- &- \\
RGB-N \cite{Zhou2018LearningRF} &0.364 &0.408 &0.334 &0.379 &0.261&0.383 &0.290 & 0.424 \\
EXIF-consistency \cite{Huh2018FightingFN} &0.127 &0.204 &0.102 &0.276 & 0.420 &0.520 & 0.415 &0.504 \\
GSR-Net \cite{Zhou2020GenerateSA} &0.553 &0.574 &0.439 & 0.489 &0.462 & 0.525 &0.446 &0.555 \\
Encoder-Decoder (DeepLabV3+) \cite{Chen2018EncoderDecoderWA} &0.505 &0.483 &0.374 & 0.415 &0.401 & 0.468 &0.365 &0.505 \\
\hline
\textbf{TBNet} (ours) &\textbf{0.615} &\textbf{0.597} &\textbf{0.456} &\textbf{0.518} &\textbf{0.503} &\textbf{0.584} &\textbf{0.454} &\textbf{0.567} \\
\hline
\end{tabular}
\vspace{-1.0em}
\end{table*}

\subsection {Performance Evaluation and Comparisons}
We first assess the performance of the TBNet on four public image manipulation localization datasets and then compare it with those of the competing state-of-the-art approaches. Among these approaches, if the codes can be obtained, the ImageNet is first used to pre-train them, and then 100,000 training samples are further utilized to fine-tune the models, and finally, four public image manipulation localization datasets are employed to assess their performance. If the codes are not available, the results reported by the corresponding references are used. Moreover, for a fair comparison, if the performance of a model trained by our training samples is lower than that of the corresponding reference, the results reported by the corresponding references are also used. Their results are shown in Table I.
An examination of the results presented in Table I reveals the following observations:

1) TBNet shows the best performance regardless of the dataset and approach used, and a large improvement in MCC and F1 is obtained compared with state-of-the-art algorithms. For example, when the CASIA1.0 dataset is employed, MCC and F1 of TBNet reach 0.615 and 0.597, respectively, but compared to the baseline, the corresponding improvements can achieve 11\% and 11.4\%, respectively. Similarly, when the Carvalho dataset is utilized, the corresponding improvements can reach 10.2\% and 11.6\%, respectively. Moreover, when the In-The-Wild dataset is utilized, the corresponding improvements reach 8.9\% and 6.2\%, respectively. The performance of the TBNet is much better than that of the DeepLabV3+ because the frequency information is fully used, and the ACF module is proposed to adaptively fuse the RGB feature and the frequency feature, and finally, the BAL module is used to locate the boundary artifacts, and the parameters of the RGB stream, the frequency stream, and the boundary artifacts location network are jointly optimized. Thus, TBNet experimentally exhibits very good generalization ability, and these experimental results prove the effectiveness and robustness of the TBNet.

2) In these approaches, the GSR-Net obtains second place when different datasets are used, proving that GSR-Net is highly effective. However, GSR-Net uses a complex procedure that consists of the generation, segmentation, and refinement stages. In the generation stage, more samples that are difficult to be distinguished are generated by the GAN networks, and then these samples are used to enhance the recognition of the segmentation network, and finally, the refinement stage further guides the network to pay attention to the boundary. In TBNet, to obtain much more training samples, the operation of overlap-slipping is randomly performed on the holistic images of the CASIA2.0 dataset, and then the network is optimized using these training samples. Thus, the procedure is much simpler than that of the GSR-Net. Moreover, frequency information is ignored in GST-Net; thus, its performance is worse than that of the TBNet. For example, when the CASIA1.0 dataset is used, the MCCs of the TBNet and the GSR-Net are 0.615 and 0.553, respectively, with the improvement reaching 6.2\%. Similarly, when the Carvalho dataset is employed, the MCCs achieved by TBNet and GSR-Net are 0.503 and 0.462, respectively, reaching an improvement of 4.1\%.

3) RGB-N transforms the tampering location into an object detection task, and the RGB stream and the noise stream are combined by bilinear pooling where the two-stream network architecture is used, and the noise stream is obtained by the famous SRM \cite{Fridrich2012RichMF} to make the image high-frequency information-sensitive. In TBNet, the two-stream network architecture is also employed, but the frequency stream (the noise stream is used in the RGB-N) is utilized, and the boundary artifacts location network is also embedded into TBNet. The SRM is highly beneficial for distinguishing tampered areas from real areas and is only sensitive to high-frequency information, but in the TBNet, the AFS module is proposed to adaptively select the appropriate frequency information to mine inconsistent statistics and eliminate the interference of redundant statistics. Thus, the performance of the TBNet is much better than that of RGB-N. For example, assessed on the COVER dataset, the MCC and F1 values of TBNet are 0.456 and 0.518, respectively, and the MCC and F1 values of RGB-N are 0.334 and 0.379, corresponding to improvements of 12.2\% and 13.9\%, respectively.

4) MFCN first predicts the edge mask and the segmentation mask, and then the intersection of the boundary mask after the hole filling and the segmentation mask is taken as the tampered area, but the latent relationship between the boundary mask and the segmentation mask needs to be further mined. In TBNet, the boundary artifacts location network is also used, but the parameters of RGB stream, frequency stream, and the boundary artifacts location network are jointly optimized, and their latent complementary relationships are well-mined. When the CASIA dataset is used, the MCCs of the TBNet and the MFCN are 0.615 and 0.520, respectively, and the improvement can reach 9.5\%. Thus, the performance of TBNet is clearly superior to that of MFCN. For EXIF-consistency, the metadata is used to learn the features which are useful for the manipulation localization, but the robustness and the discrimination of the learned feature need to be further improved. NOI and the CFA tried different ideas in the early few years and provided much useful guidance, but their performance needs to be further improved compared with TBNet.

\section{Ablation Study}
An ablation study was performed using the TBNet model to analyze the contribution of each component. In this investigation, four aspects were considered: 1) effectiveness of the AFS module, 2) benefits of the ACF and BAL modules, 3) robustness to different attacks, and 4) qualitative visualization. In the following, we discuss these four aspects separately.

\subsection {Effectiveness of the AFS module}

In this section, we will assess the effectiveness of the AFS module. The frequency information was often ignored in the early few years, but it may contain key image tampering clues. In recent years, the frequency information has attracted research attention, but only the high-frequency information is often used, and the low-frequency information is simply discarded. Thus, the AFS module is designed to adaptively select the appropriate frequency information to mine inconsistent statistics and eliminate the interference of redundant statistics. We assess the effectiveness of the AFS module on the CASIA1.0 dataset that is the largest of these four datasets and includes splicing and copy-move. Their results are given in Table II. We note that in the frequency stream, the YCbCr color space is utilized, and each component has 64 frequency channels. 'Cb+Cr' indicates the results when the components of 'Cb' and 'Cr' are jointly used, but all 128 frequency channels are equally utilized. 'Cb+Cr (AFS)' denotes the results when the components of 'Cb' and 'Cr' are jointly used, but all 128 frequency channels are adaptively selected by the AFS module. 'Y+Cb+Cr' and 'Y+Cb+Cr (AFS)' have similar meanings with 'Cb+Cr' and 'Cb+Cr (AFS)'. Moreover, 'Dim' indicates the number of frequency channels. An examination of this table shows the following:

1) When each component of 'Y', 'Cb', and 'Cr' is used individually, their MCC values are 0.533, 0.577, and 0.567, respectively, and the worst performance is found for 'Y'. When the components of 'Y', 'Cb' and 'Cr' are jointly used, the MCCs of 'Cb+Cr' and 'Y+Cb+Cr' are 0.589 and 0.578, respectively. Their performance is improved slightly compared with each component of 'Y', 'Cb', and 'Cr'; thus, different components are complementary to some extent.

2) However, we also find when 'Y' is further used in the 'Cb+Cr', its performance is slightly decreased, and the 'Y' component does not appear to be suitable for image tampering detection compared with the ’Cb' and 'Cr' components \cite{Wang2009EffectiveIS}; however, this is not true. The reason for this is that there is a large amount of redundant information in the 'Y' component to interfere with the judgment, but we believe that each component contains some channels that are highly beneficial for detecting the tampered image, and these components are complementary. It is vital to adaptively select the appropriate frequency information. If we directly concatenate these data and each frequency is equally treated, the interference of redundant frequency is included. Therefore, we proposed the AFS module to mine inconsistent statistics and eliminate the interference of redundant statistics. When the AFS module is utilized, the performance of 'Cb+Cr' and 'Y+Cb+Cr' can be improved. For example, the F1s of 'Cb+Cr' and ‘Cb+Cr (AFS)' are 0.558 and 0.578, respectively, and the improvement can reach 2\%. Similarly, MCC and F1 of the 'Y+Cb+Cr' are 0.615 and 0.597, respectively, and MCC and F1 of the 'Y+Cb+Cr (AFS)' are 0.578 and 0.543, respectively, so that the improvements of 'Y+Cb+Cr (AFS)' of 3.7\% (MCC) and 5.4\% (F1), respectively, is achieved. This proves the AFS module is highly effective.

3) To further demonstrate the mechanism of the effect of the AFS module, heatmaps of each frequency channel in the 'Y+Cb+Cr (AFS)' are given in Fig. 5. It is observed that the weights of some frequency channels are very small, but some frequency channels have large weights. Moreover, the selected frequency channels with deeper color are from different components including 'Y', 'Cb', and 'Cr'. This further proves that the ’Y’ component is also useful, and these components are complementary.
\begin{table}[h]
\fontsize{10}{10}\selectfont
\renewcommand{\arraystretch}{1.5}
\caption{Effectiveness of the AFS module on CASIA1.0}
\centering
\begin{tabular}{l|c|c|c}
\hline
Components &Dim &MCC &F1\\
\hline
Y	&64&0.533 &0.524\\
Cb &64&0.577 &0.542\\
Cr &64&0.567 &0.547\\
Cb+Cr &128&0.589 &0.558\\
Cb+Cr (AFS) &128 &0.593 &0.578\\
Y+Cb+Cr &192 &0.578 &0.543\\
Y+Cb+Cr (AFS) &192 &0.615 &0.597 \\
\hline
\end{tabular}
\vspace{-1.0em}
\end{table}

\begin{figure*}[htb]
\begin{center}
\includegraphics[width=6in,height = 1.8in]{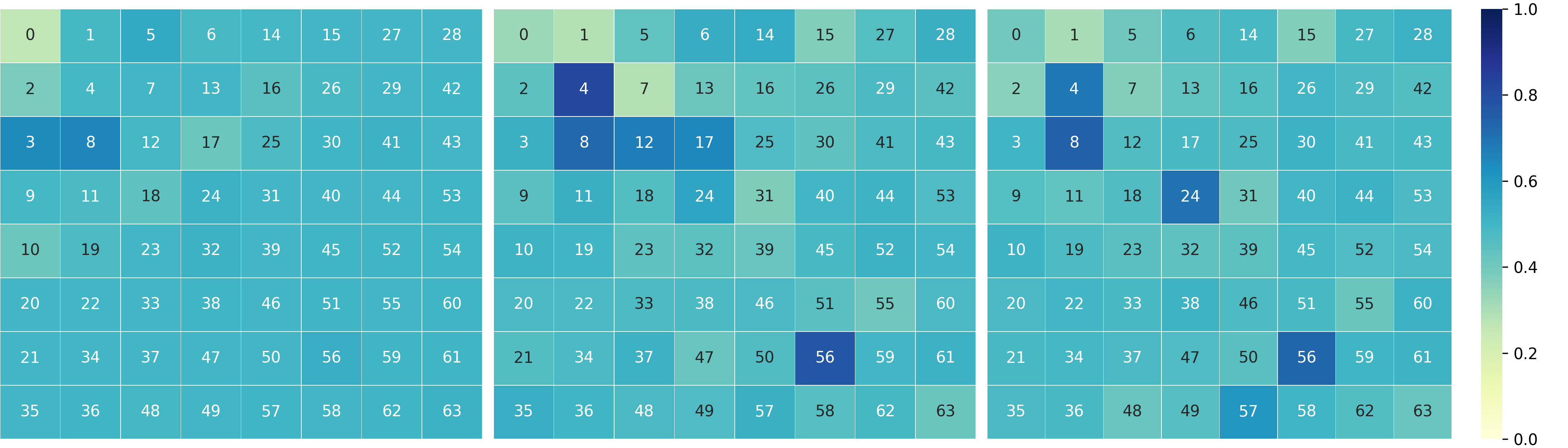}
\vspace{-0.5em}
\caption{Heatmaps for each channel in the YCbCr color space where 64 frequency channels are used to represent each component, and the number is the index with the range from 0 to 63. The AFS module is used to adaptively assign the weight for each frequency, and the deeper color corresponds to greater weight. From left to right, these heatmaps are from 'Y', 'Cb' and 'Cr'}
\end{center}
\vspace{-0.5em}
\end{figure*}
\begin{table*}
\centering
\fontsize{10}{10}\selectfont
\renewcommand{\arraystretch}{1.6}
\caption{Advantage of the ACF and BAL modules}
\centering
\begin{tabular}{ccccccccc}
\hline
\multirow{1}{*}{Datasets} & \multicolumn{2}{c} {CASIA1.0}& \multicolumn{2}{c} {COVER}& \multicolumn{2}{c} {Carvalho}& \multicolumn{2}{c} {In-The-Wild} \\
\cline{1-9}
Metrics &MCC &F1 &MCC &F1 &MCC &F1 &MCC &F1\\
\hline
RNet &0.505 & 0.483 &0.374 & 0.415 & 0.401&0.468 & 0.365 &0.505 \\
FNet &0.529&0.474 &0.403&0.389 &0.432&0.474 &0.391&0.516 \\
\hline
TNet &0.579&0.566 &0.437&0.466 & 0.482&0.497 & 0.438&0.538\\
TNet+ACF &0.594&0.571 & 0.443&0.472 &0.498&0.511 & 0.441&0.547\\
\hline
RNet+BAL &0.533&0.512& 0.394&0.437 &0.436&0.509 &0.386&0.522 \\
FNet+BAL &0.535&0.471 &0.419&0.390 & 0.441&0.468 & 0.389&0.519\\
\hline
TNet+BAL &0.609&0.587 &0.451&0.498 & 0.489&0.563 & 0.439&0.541\\
TBNet &0.615&0.597 &0.456&0.518 &0.503&0.584 &0.454&0.567 \\
\hline
\end{tabular}
\end{table*}

\subsection {Benefits of the ACF and BAL modules}
In this section, we assess the benefits of the ACF and BAL modules. First, the RGB stream (the network is the same as DeepLabV3+, and we call it as RNet) and the frequency stream (the network architecture is the same as for the RGB stream, but the frequency information \cite{Xu2020LearningIT} is fed into it, and we name it as FNet) of the TBNet are separately evaluated, and then these two streams are fused by the simple feature concatenation (TNet) and the ACF module (TNet+ACF), respectively. Moreover, the boundary artifact location module is further embedded into the RNet, FNet, and TNet, and we separately call them the RNet+BAL, FNet+BAL, and TNet+BAL, respectively. Finally, the two-stream network, the ACF module, and the BAL module are jointly employed (TBNet). Their results are shown in Table III, and we can observe the following:

1) When only the frequency information is employed, the MCC of the FNet is slightly better than that of the RNet regardless of which dataset is used, and the F1 of the FNet is also comparable to the RNet. For example, when the CASIA1.0 dataset is utilized, the MCC values of RNet and FNet are 0.505 and 0.529, respectively, for which the improvement can reach 2.4\%. Similarly, when the In-The-Wild dataset is employed, the MCC values of RNet and FNet are 0.365 and 0.391, respectively, showing an improvement of 2.6\%. Moreover, when these two streams are fused, their performance can obtain a strong improvement. For example, when the CASIA1.0 dataset is utilized, the MCC and F1 values of TNet are 0.579 and 0.566, respectively. Compared to RNet and FNet, the improvement reaches 7.4\% (RNet, MCC), 5\% (FNet, MCC), 8.3\% (RNet, F1), and 9.2\% (FNet, F1), respectively; thus, these results prove that the frequency information is highly useful for detecting the tampered image, and the RGB information and the frequency information are complementary.

2) In TNet, the simple feature concatenation is used, but the latent complementary relationship between the RGB stream and the frequency stream needs to be further mined so that the ACF module using the cross-attention fusion scheme is proposed to adaptively fuse the RGB feature and the frequency feature. Thus, the latent complementary relationship between these two streams is well explored, and the performance of the latter is slightly better than that of the former. For example, when the Carvalho dataset is used, the MCC values of TNet+ACF and TNet are 0.498 and 0.482, respectively, and the improvement reaches 1.6\%. We also reach similar conclusions for the other datasets. Thus, these results demonstrate the effectiveness of the ACF module.

3) The boundary artifact location is also highly beneficial for the tampered image detection, and when it is embedded into RNet, and TNet, their performance can be further improved. This is because the BAL module and the RNet (or TNet) are jointly optimized, and their latent complementary relationships are well-mined. For example, when the CASIA1.0 dataset is utilized, the MCCs of the RNet+BAL, TNet+BAL, RNet, and TNet are 0.533, 0.609, 0.505, and 0.579, respectively, and the improvement can reach 2.8\% (RNet) and 3.0\% (TNet). We also observe similar results for other datasets. We note that when the BAL module is directly embedded into the FNet, the improvement of the FNet+BAL is quite limited. Finally, the ACF module and the BAL module are embedded into the TNet, and its performance is further improved. For example, when the Carvalho dataset is used, the MCC and F1 values of TBNet are 0.503 and 0.584, respectively. Compared with TNet+BAL, an improvement of 1.4\% (MCC) and 2.1\% (F1) was achieved. Thus, this proves that the BAL module is quite effective, and it is complementary to the RGB stream and the frequency stream.

\subsection{Robustness to Attacks}
In this section, to further assess the robustness of the TBNet, different attacks including JPEG compression and scaling, are first used to the testing images from the CASIA1.0 and Carvalho datasets, respectively, and then the performance of TBNet is assessed on these two datasets. Their results are shown in Figs. 6 and 7. We note that in the scaling attacks, the scaling ratios of 0.7 and 0.5 are used in our experiments, and the JPEG compression consists of quality factors of 70 and 50. $MCC\_CASIA$ and $MCC\_carvalho$ indicate the MCC performance of the TBNet when the CASIA1.0 and Carvalho datasets are separately employed. $F1\_CASIA$ and $F1\_carvalho$ indicate the F1 performance of the TBNet when the CASIA1.0 and Carvalho datasets are separately used. It is observed from Fig. 6 that the MCC and F1 lines are essentially a straight line when the scaling attack is used so that the scaling attack has almost no effect on the TBNet. Similarly, when the JPEG compression attack is utilized in Fig. 7, the MCC and F1 lines show a slight downward trend, indicating that this attack degrades the performance of TBNet slightly, but it still can mine the remaining tampering traces from all of the channels. We think this is because some high-frequency information is lost in the jpeg compression process, and this information is very important for detecting the tampering image. Thus, these experimental results prove the stability of TBNet.
\begin{figure}[htbp]
\begin{center}
\includegraphics[width=3.in,height = 2.in]{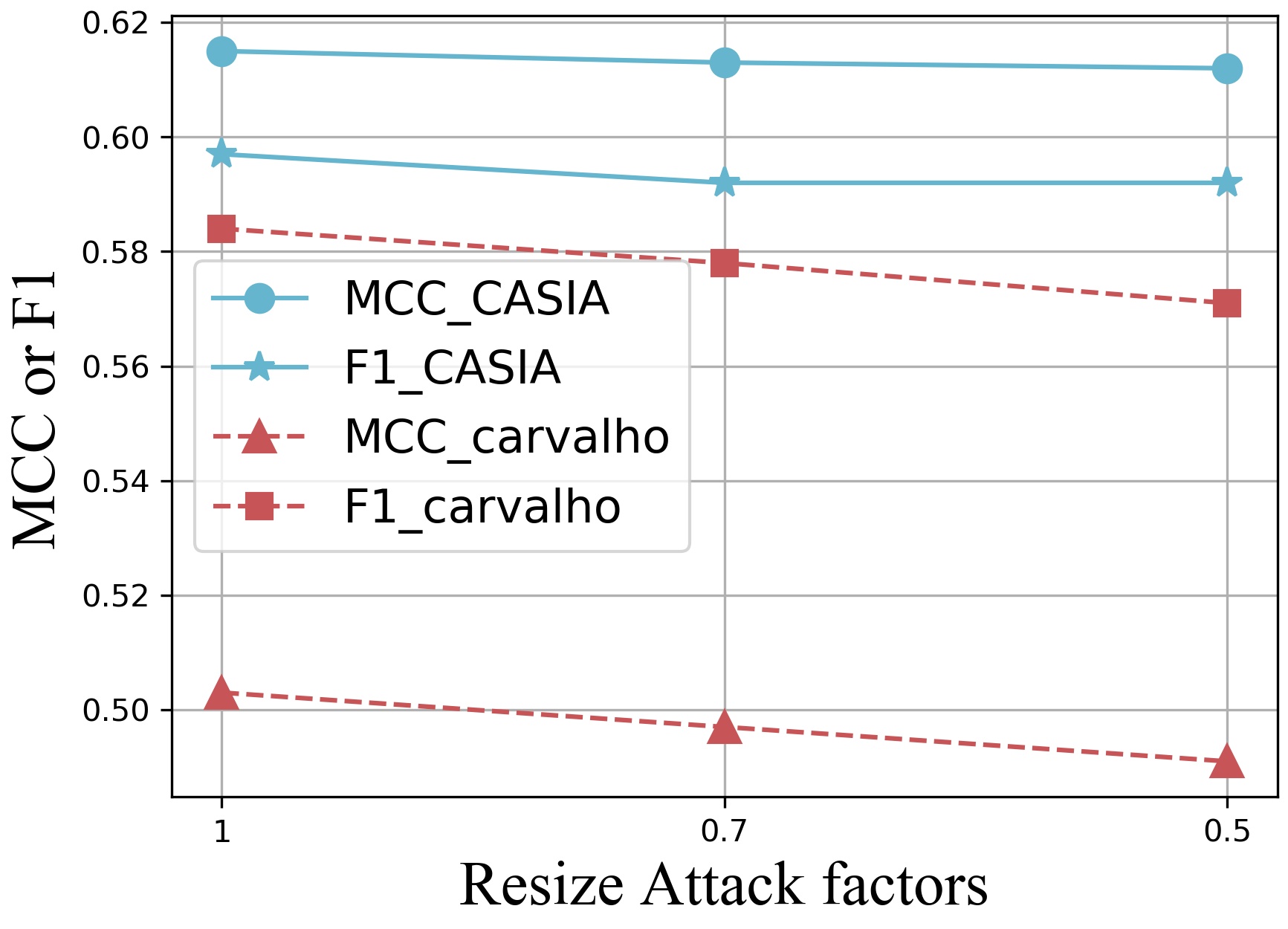}
\vspace{-0.5em}
\caption{Analysis of robustness under the resizing attacks on the CASIA1.0 and Carvalho datasets where the scale attacks use the scaling ratios of 0.7 and 0.5}
\vspace{-1.5em}
\end{center}
\end{figure}

\begin{figure}[htbp]
\begin{center}
\includegraphics[width=3.in,height = 2.in]{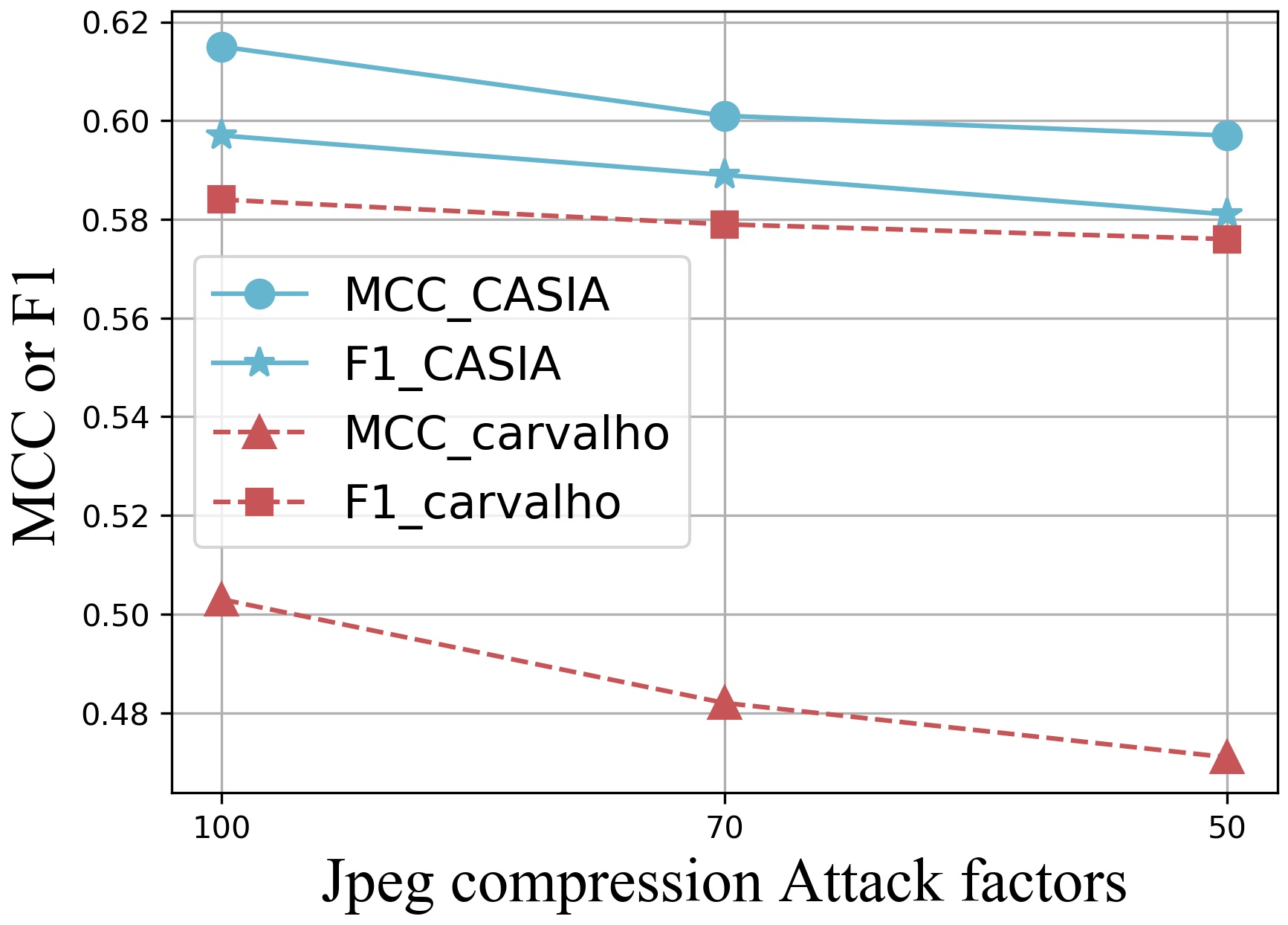}
\vspace{-0.5em}
\caption{Analysis of robustness under the JPEG compression attacks on the CASIA1.0 and Carvalho datasets where the JPEG compression consists of quality factors of 70 and 50.}
\vspace{-1.5em}
\end{center}
\end{figure}

\subsection{Qualitative visualization}
To further prove the effectiveness and robustness of TBNet, in this section, some visualizations are given. In our experiments, three aspects are visualized: 1) Importance of the frequency stream, 2) Complementarity between the image manipulation regions and the boundary artifacts of the TbNet, 3) Comparison with state-of-the-art methods. We note that these images are randomly selected from four image manipulation localization datasets, but the manipulation operations of splicing and copy-moving are included. The results are shown in Figs. 8-10, respectively. The following can be observed from these figures:
\begin{figure}[htb]
\begin{center}
\vspace{-1.0em}
\includegraphics[width=3.5in,height = 2.5in]{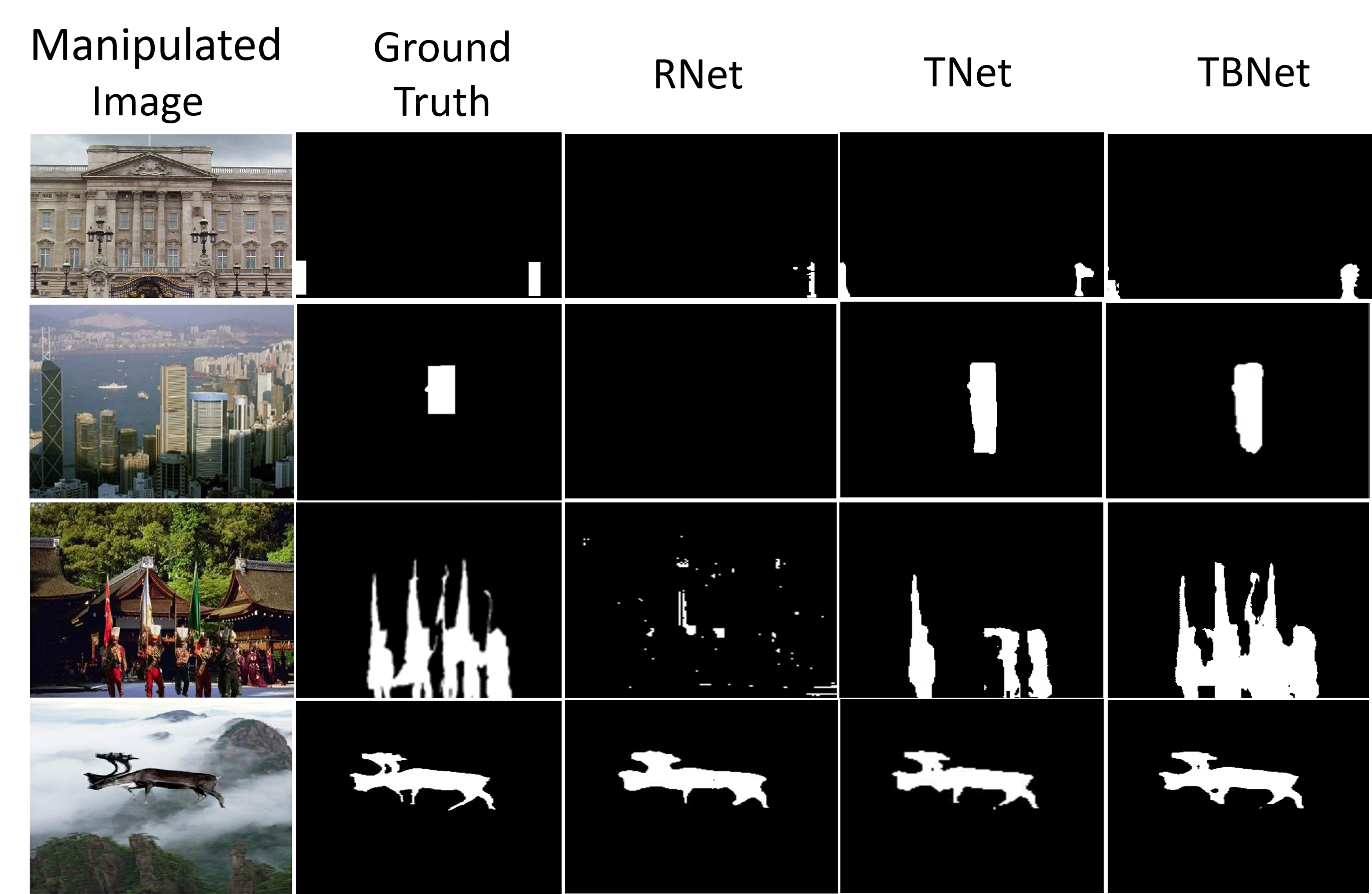}
\caption{Qualitative visualization of RNet, TNet, and TBNet. The manipulation of the images in the top two rows are  carried out with copy-move, and the manipulation of other images is carried out by splicing. }
\vspace{-2.5em}
\end{center}
\end{figure}
\begin{figure}[htb]
\begin{center}
\includegraphics[width=3.5in,height = 2.5in]{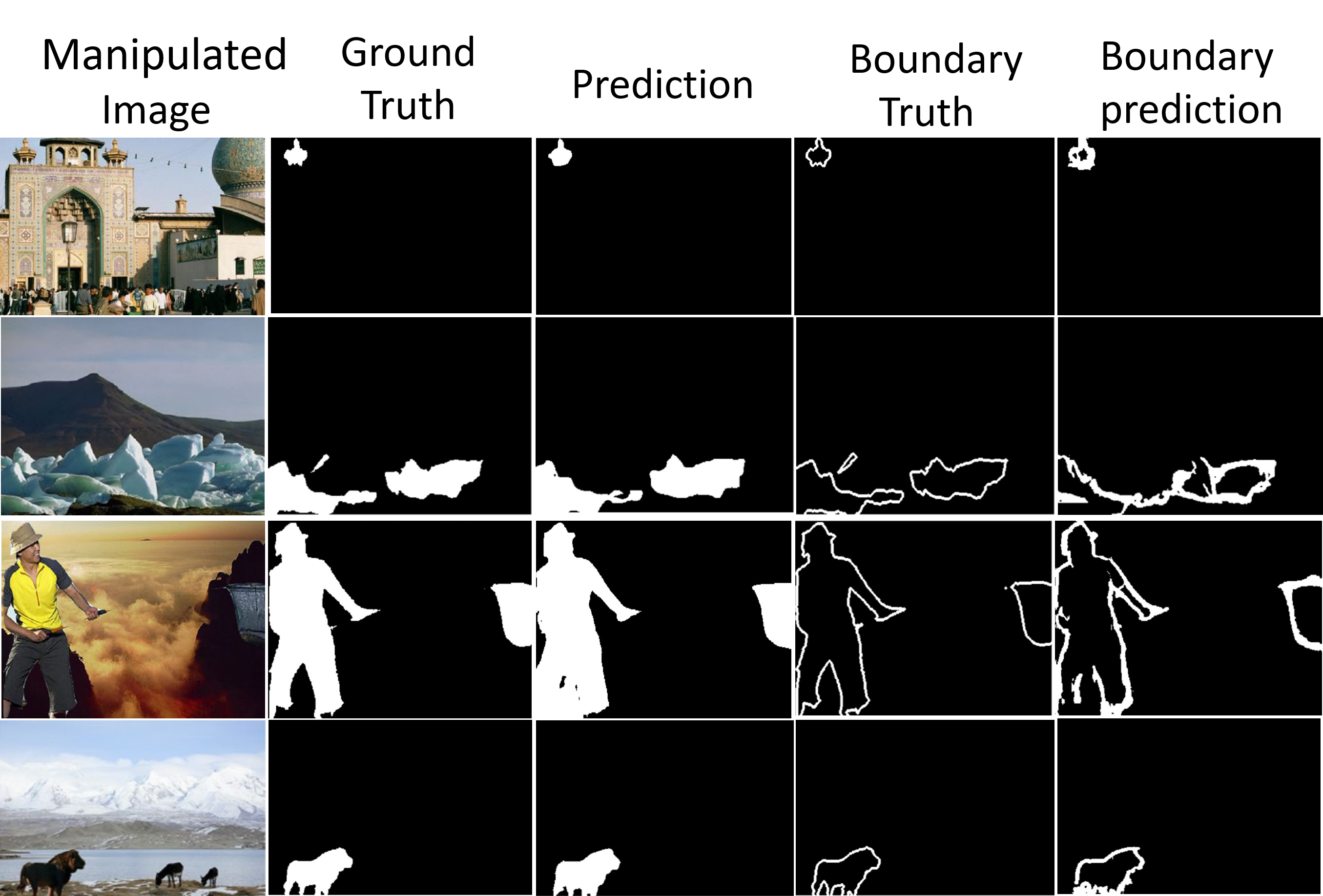}
\caption{Qualitative visualization of the image manipulation regions and the boundary artifacts when TBNet is employed. The manipulation of images in the top two rows is carried out with copy-move, and the manipulation of other images is carried out by splicing.}
\vspace{-1.em}
\end{center}
\end{figure}

\begin{figure*}[htb]
\begin{center}
\includegraphics[width=7in,height = 5.3in]{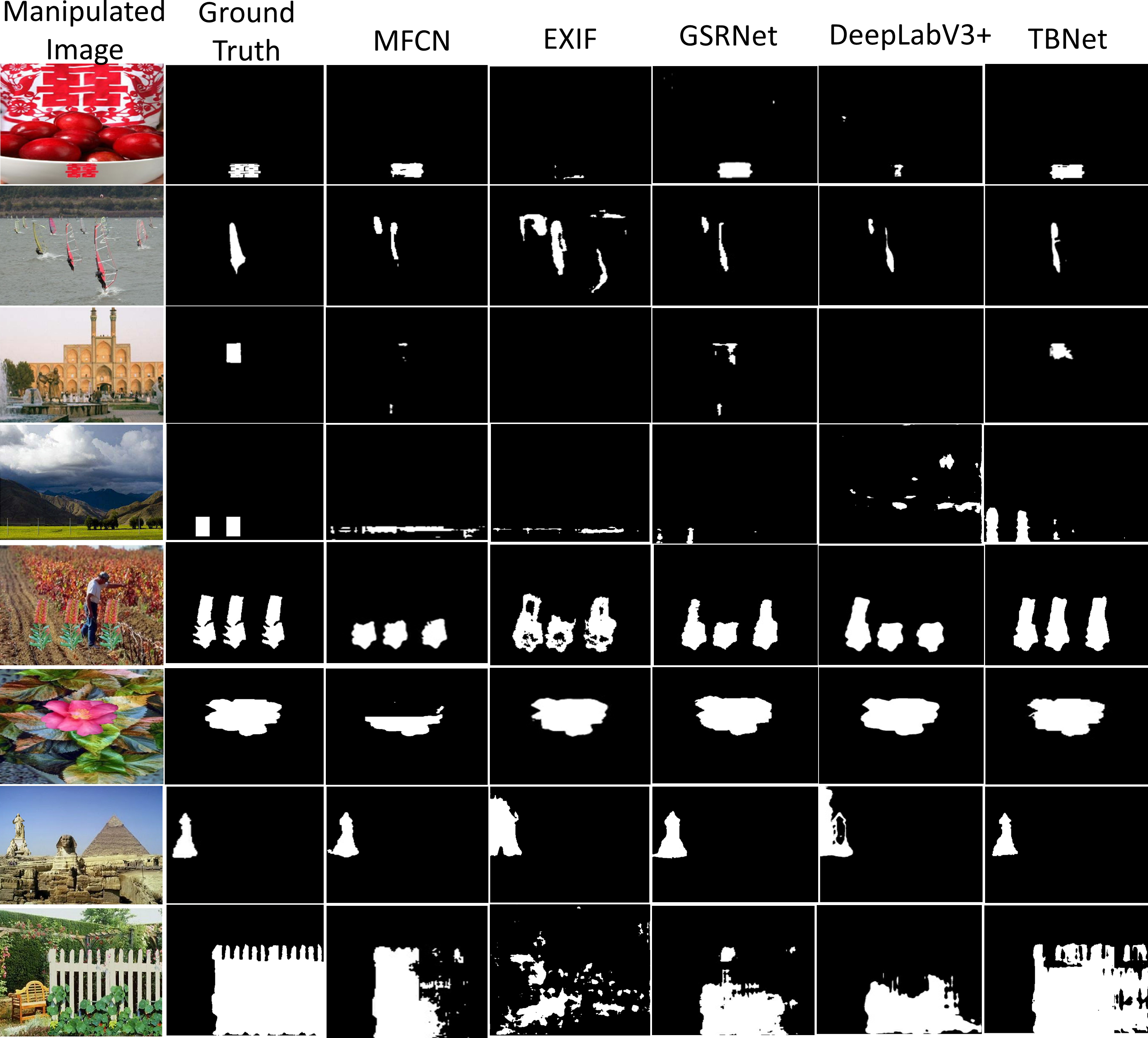}
\caption{Qualitative results of different location algorithms. The first column shows manipulated images on different datasets. The second column is the ground truth. From the third to sixth columns, they indicate the final manipulation segmentation prediction of the MFCN, EXIF, GSRNet, and DeepLabV3+, respectively. The last column is the results of TBNet. We note that the manipulation of images in the top four rows is carried out with copy-move, and the manipulation of other rows is carried out by splicing.}
\vspace{-1.5em}
\end{center}
\end{figure*}

1) As shown in Fig. 8, when only RNet is used, the tampering regions of the top three columns cannot be detected. However, when it is combined with the frequency stream by simple feature concatenation (TNet), these tampering regions are better detected; this indicates that the frequency information is important, and is beneficial for finding tampering clues; moreover, the RGB stream and the frequency stream are complementary. Additionally, it is observed from the third row in Fig. 8 that when TNet is used, some tampering regions are ignored, but when TBNet is employed, all of the tampering regions can be well detected. This is because the AFS, ACF, and BAL modules are jointly employed, and their complementary relationships are further mined.

2) Fig. 9 shows the image manipulation prediction and the boundary prediction of TBNet.  It is observed from the figure  that the image manipulation prediction and the boundary prediction of TbNet are complementary. For example, the regions between two legs are detected as the tampering regions in the manipulation prediction of the third column, but from the ground truth, we can find that this is not true. Fortunately, the boundary prediction of the TBNet can well detect the boundary, and this can compensate for the shortcomings of the image manipulation prediction.

3) Fig. 10 shows the visualizations of TBNet and other state-of-the-art methods. It is observed that TBNet outperforms other state-of-the-art methods regardless of which manipulation is used. For example, in the second row (the copy-move manipulation), from the ground truth, we know that only the sailboat is the tampering region, and TBNet can well detect this region, but more than one sailboat are detected by other state-of-the-art methods. Similarly, in the last row (the splicing manipulation), the tampering region is the whole railings, and TBNet can locate almost all areas of the railings, but many areas are ignored when other state-of-the-art methods are used. Thus, TBNet can locate the tampered area more accurately, divide the edge more finely, and its results are closer to the real label.

\section{Conclusion}
This paper proposes a novel end-to-end TBNet algorithm for generic image manipulation localization in which the RGB stream, the frequency stream, and the boundary artifacts location are jointly explored in a unified framework. Moreover, we design an AFS module to adaptively select the appropriate frequency to mine inconsistent statistics and eliminate the interference of redundant statistics. Additionally, we propose an ACF module to adaptively fuse the RGB feature and the frequency feature. Finally, we embed the boundary artifacts location network into the two-stream encoder to locate the boundary artifacts the parameters of which are jointly updated. The results of extensive experiments conducted on four public benchmarks of image manipulation localization datasets demonstrate that TBNet can significantly outperform state-of-the-art image manipulation localization methods in terms of both MCC and F1. The ablation study also proves that the frequency information is highly useful for finding the tampering clues, and is complementary to the RGB information. The AFS module is highly effective for adaptively selecting the appropriate frequency, and the ACF module can well fuse the RGB feature and the frequency feature. Moreover, the BAL module is also complementary to the two-stream encoder.

In the future, we intend to explore a more effective scheme for adaptively selecting the frequency information, and design a novel and effective feature fusion module.
\ifCLASSOPTIONcaptionsoff
  \newpage
\fi

\normalem
\bibliographystyle{IEEEtran}
 \bibliography{TBNET}

\begin{thebibliography}{10}
\providecommand{\url}[1]{#1}
\csname url@samestyle\endcsname
\providecommand{\newblock}{\relax}
\providecommand{\bibinfo}[2]{#2}
\providecommand{\BIBentrySTDinterwordspacing}{\spaceskip=0pt\relax}
\providecommand{\BIBentryALTinterwordstretchfactor}{4}
\providecommand{\BIBentryALTinterwordspacing}{\spaceskip=\fontdimen2\font plus
\BIBentryALTinterwordstretchfactor\fontdimen3\font minus
  \fontdimen4\font\relax}
\providecommand{\BIBforeignlanguage}[2]{{%
\expandafter\ifx\csname l@#1\endcsname\relax
\typeout{** WARNING: IEEEtran.bst: No hyphenation pattern has been}%
\typeout{** loaded for the language `#1'. Using the pattern for}%
\typeout{** the default language instead.}%
\else
\language=\csname l@#1\endcsname
\fi
#2}}
\providecommand{\BIBdecl}{\relax}
\BIBdecl

\bibitem{Bunk2017DetectionAL}
J.~Bunk, J.~H. Bappy, and T.~M. et~al., ``Detection and localization of image
  forgeries using resampling features and deep learning,'' in \emph{IEEE
  Conference on Computer Vision and Pattern Recognition Workshops}, 2017, pp.
  1881--1889.

\bibitem{Park2018DoubleJD}
J.-S. Park, D.~Cho, W.~Ahn, and H.~Lee, ``Double jpeg detection in mixed jpeg
  quality factors using deep convolutional neural network,'' in \emph{Computer
  Vision - {ECCV} 2018 - 15th European Conference, Munich, Germany, September
  8-14}, 2018.

\bibitem{Bappy2019Hybrid}
J.~H. Bappy, C.~Simons, L.~Nataraj, B.~S. Manjunath, and A.~K. Roy{-}Chowdhury,
  ``Hybrid {LSTM} and encoder-decoder architecture for detection of image
  forgeries,'' \emph{{IEEE} Trans. Image Process.}, vol.~28, no.~7, pp.
  3286--3300, 2019.

\bibitem{Bappy2017ExploitingSS}
J.~H. Bappy, A.~Roy-Chowdhury, J.~Bunk, L.~Nataraj, and B.~S. Manjunath,
  ``Exploiting spatial structure for localizing manipulated image regions,'' in
  \emph{IEEE International Conference on Computer Vision}, 2017, pp.
  4980--4989.

\bibitem{Mohammed2018BoostingIF}
T.~Mohammed, J.~Bunk, and L.~N. et~al., ``Boosting image forgery detection
  using resampling features and copy-move analysis,'' vol. abs/1802.03154,
  2018.

\bibitem{Zhou2020GenerateSA}
P.~Zhou, B.-C. Chen, X.~Han, M.~Najibi, and L.~Davis, ``Generate, segment and
  replace: Towards generic manipulation segmentation,'' in \emph{The
  Thirty-Fourth Conference on Artificial Intelligence, NY, USA, February 7-12},
  2020, pp. 13\,058--13\,065.

\bibitem{Zhou2018LearningRF}
P.~Zhou, X.~Han, V.~Morariu, and L.~Davis, ``Learning rich features for image
  manipulation detection,'' in \emph{IEEE/CVF Conference on Computer Vision and
  Pattern Recognition}, 2018, pp. 1053--1061.

\bibitem{Fridrich2012RichMF}
J.~J. Fridrich and J.~Kodovsk{\'{y}}, ``Rich models for steganalysis of digital
  images,'' \emph{{IEEE} Trans. Inf. Forensics Secur.}, vol.~7, no.~3, pp.
  868--882, 2012.

\bibitem{Cozzolino2015SplicebusterAN}
D.~Cozzolino, G.~Poggi, and L.~Verdoliva, ``Splicebuster: {A} new blind image
  splicing detector,'' in \emph{IEEE}.

\bibitem{Cozzolino2017RecastingRL}
D.~Cozzlino, G.~Poggi, and L.~Verdoliva, ``Recasting residual-based local
  descriptors as convolutional neural networks: an application to image forgery
  detection,'' in \emph{Proceedings of the 5th ACM Workshop on Information
  Hiding and Multimedia Security}, 2017, pp. 159--164.

\bibitem{Rao2016ADL}
Y.~Rao and J.~Ni, ``A deep learning approach to detection of splicing and
  copy-move forgeries in images,'' in \emph{IEEE International Workshop on
  Information Forensics and Security}, 2016, pp. 1--6.

\bibitem{Liu2018ImageFL}
Y.~Liu, Q.~Guan, X.~Zhao, and Y.~Cao, ``Image forgery localization based on
  multi-scale convolutional neural networks,'' in \emph{Proceedings of the 6th
  {ACM} Workshop on Information Hiding and Multimedia Security, Innsbruck,
  Austria, June 20-22}, 2018, pp. 85--90.

\bibitem{Wu2018ImageCF}
Y.~Wu, W.~Abd-Almageed, and P.~Natarajan, ``Image copy-move forgery detection
  via an end-to-end deep neural network,'' in \emph{IEEE Winter Conference on
  Applications of Computer Vision}, 2018, pp. 1907--1915.

\bibitem{Wu2018BusterNetDC}
Y.~Wu, W.~Abd{-}Almageed, and P.~Natarajan, ``Busternet: Detecting copy-move
  image forgery with source/target localization,'' in \emph{Computer Vision -
  {ECCV} 2018 - 15th European Conference, Munich, Germany, September 8-14},
  2018, pp. 170--186.

\bibitem{Islam2020DOAGANDA}
A.~Islam, C.~Long, A.~Basharat, and A.~Hoogs, ``{DOA-GAN:} dual-order attentive
  generative adversarial network for image copy-move forgery detection and
  localization,'' in \emph{IEEE Conference on Computer Vision and Pattern
  Recognition, Seattle, WA, USA, June 13-19}, 2020, pp. 4675--4684.

\bibitem{Liu2018DeepFN}
B.~Liu and C.~Pun, ``Deep fusion network for splicing forgery localization,''
  in \emph{Computer Vision - {ECCV} 2018 Workshops - Munich, Germany, September
  8-14,}, vol. 11130, 2018, pp. 237--251.

\bibitem{Bi2019RRUNetTR}
X.~Bi, Y.~Wei, B.~Xiao, and W.~Li, ``Rru-net: The ringed residual u-net for
  image splicing forgery detection,'' in \emph{IEEE/CVF Conference on Computer
  Vision and Pattern Recognition Workshops}, 2019, pp. 30--39.

\bibitem{Xiao2020ImageSF}
B.~Xiao, Y.~Wei, X.~Bi, W.~Li, and J.~Ma, ``Image splicing forgery detection
  combining coarse to refined convolutional neural network and adaptive
  clustering,'' \emph{Inf. Sci.}, vol. 511, pp. 172--191, 2020.

\bibitem{Liu2020ExposingSF}
B.~Liu and C.-M. Pun, ``Exposing splicing forgery in realistic scenes using
  deep fusion network,'' \emph{Inf. Sci.}, vol. 526, pp. 133--150, 2020.

\bibitem{Zhong2020AnED}
J.~Zhong and C.-M. Pun, ``An end-to-end dense-inceptionnet for image copy-move
  forgery detection,'' \emph{IEEE Transactions on Information Forensics and
  Security}, vol.~15, pp. 2134--2146, 2020.

\bibitem{Barni2021CopyMS}
M.~Barni, Q.-T. Phan, and B.~Tondi, ``Copy move source-target disambiguation
  through multi-branch cnns,'' \emph{IEEE Transactions on Information Forensics
  and Security}, vol.~16, pp. 1825--1840, 2021.

\bibitem{salloum2018image}
R.~Salloum, Y.~Ren, and C.-C.~J. Kuo, ``Image splicing localization using a
  multi-task fully convolutional network (mfcn),'' \emph{Journal of Visual
  Communication and Image Representation}, vol.~51, pp. 201--209, 2018.

\bibitem{Chen2018EncoderDecoderWA}
L.~Chen, Y.~Zhu, and G.~P. et~al., ``Encoder-decoder with atrous separable
  convolution for semantic image segmentation,'' in \emph{Computer Vision -
  {ECCV} 2018 - 15th European Conference, Munich, Germany, September 8-14},
  vol. 11211, 2018, pp. 833--851.

\bibitem{Ahmed1974Discrete}
N.~Ahmed, T.~R. Natarajan, and K.~R. Rao, ``Discrete cosine transform,''
  \emph{{IEEE} Trans. Computers}, vol.~23, no.~1, pp. 90--93, 1974.

\bibitem{Hu2018SqueezeandExcitationN}
J.~Hu, L.~Shen, and G.~Sun, ``Squeeze-and-excitation networks,'' in
  \emph{IEEE/CVF Conference on Computer Vision and Pattern Recognition}, 2018.

\bibitem{zhou2021Hierarchical}
S.~Zhou, J.~Wang, and L.~W. et~al., ``Hierarchical and interactive refinement
  network for edge-preserving salient object detection,'' \emph{{IEEE} Trans.
  Image Process.}, vol.~30, pp. 1--14, 2021.

\bibitem{Huang2020Semantic}
Z.~Huang, C.~Wang, and X.~W. et~al., ``Semantic image segmentation by
  scale-adaptive networks,'' \emph{{IEEE} Trans. Image Process.}, vol.~29, pp.
  2066--2077, 2020.

\bibitem{Han2018Robust}
J.~Han, R.~Quan, D.~Zhang, and F.~Nie, ``Robust object co-segmentation using
  background prior,'' \emph{{IEEE} Trans. Image Process.}, vol.~27, no.~4, pp.
  1639--1651, 2018.

\bibitem{Yu2017CASENet}
Z.~Yu, C.~Feng, M.~Liu, and S.~Ramalingam, ``Casenet: Deep category-aware
  semantic edge detection,'' in \emph{{IEEE} Conference on Computer Vision and
  Pattern Recognition, Honolulu, HI, USA, July 21-26}, 2017, pp. 1761--1770.

\bibitem{Dong2013CASIAIT}
J.~Dong, W.~Wang, and T.~Tan, ``Casia image tampering detection evaluation
  database,'' in \emph{International Conference on Signal and Information
  Processing}, 2013, pp. 422--426.

\bibitem{Khan2015ExposingDI}
S.~Khan and S.~S. Kulkarni, ``Exposing digital image forgeries by illumination
  color classification,'' \emph{International journal of scientific research
  and management}, vol.~3, 2015.

\bibitem{Wen2016COVERAGEA}
B.~Wen, Y.~Zhu, and S.~R. et~al., ``Coverage-a novel database for copy-move
  forgery detection,'' in \emph{IEEE International Conference on Image
  Processing}, 2016, pp. 161--165.

\bibitem{Huh2018FightingFN}
M.~Huh, A.~Liu, A.~Owens, and A.~A. Efros, ``Fighting fake news: Image splice
  detection via learned self-consistency,'' in \emph{Computer Vision - {ECCV}
  2018 - 15th European Conference, Munich, Germany, September 8-14}, 2018, pp.
  106--124.

\bibitem{Mahdian2009UsingNI}
B.~Mahdian and S.~Saic, ``Using noise inconsistencies for blind image
  forensics,'' \emph{Image Vis. Comput.}, vol.~27, pp. 1497--1503, 2009.

\bibitem{Ferrara2012ImageFL}
P.~Ferrara, T.~Bianchi, A.~Rosa, and A.~Piva, ``Image forgery localization via
  fine-grained analysis of cfa artifacts,'' \emph{IEEE Transactions on
  Information Forensics and Security}, vol.~7, pp. 1566--1577, 2012.

\bibitem{Wang2009EffectiveIS}
W.~Wang, J.~Dong, and T.~Tan, ``Effective image splicing detection based on
  image chroma,'' in \emph{16th IEEE International Conference on Image
  Processing}, 2009, pp. 1257--1260.

\bibitem{Xu2020LearningIT}
K.~Xu, M.~Qin, F.~Sun, Y.~Wang, Y.-K. Chen, and F.~Ren, ``Learning in the
  frequency domain,'' in \emph{IEEE/CVF Conference on Computer Vision and
  Pattern Recognition}, 2020, pp. 1737--1746.

\end{thebibliography}

\begin{IEEEbiography}[{\includegraphics[width=1in,height=1.25in,clip,keepaspectratio]{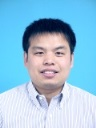}}]{Zan Gao} received his Ph.D degree from Beijing University of Posts and Telecommunications in 2011. He is currently a full Professor with the Shandong Artificial Intelligence Institute, Qilu University of Technology (Shandong Academy of Sciences). From Sep. 2009 to Sep. 2010, he worded in the School of Computer Science, Carnegie Mellon University, USA. From July 2016 to Jan 2017, he worked in the School of Computing of National University of Singapore. His research interests include artificial intelligence, multimedia analysis and retrieval, and machine learning.  He has authored over 80 scientific papers in international conferences and journals including TIP, TNNLS, TMM, TCYBE, TOMM, CVPR, ACM MM, WWW, SIGIR and AAAI, Neural Networks, and Internet of Things.
\end{IEEEbiography}

\begin{IEEEbiography}[{\includegraphics[width=0.8in,height=1in]{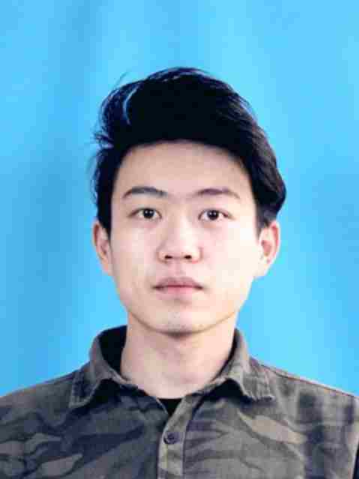}}]{Chao Sun} is pursuing his master degree in the Shandong Artificial Intelligence Institute, Qilu University of Technology (Shandong Academy of Sciences). He received his Bachelor degree from Yantai University in 2019. His research interests include artificial intelligence, multimedia analysis and retrieval, computer vision and machine learning
\end{IEEEbiography}

\begin{IEEEbiography}[{\includegraphics[width=0.8in,clip,keepaspectratio]{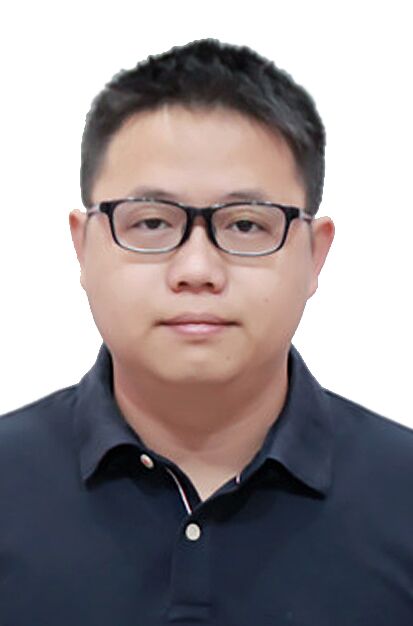}}]{Zhi-yong Cheng} is a full professor in the Shandong AI Institute, Qilu University of Technology, Jinan, China.  He received the Ph.D degree in computer science from Singapore Management University, Singapore. From 2014 to 2015, he was a visiting student with the School of Computer Science, Carnegie Mellon University, U.S.  His research interests mainly focus on large-scale multimedia content analysis and retrieval. His work has been published in a set of top forums, including ACM SIGIR, MM, WWW, IJCAI, TOIS, TKDE, TNNLS, and TCYB. 
\vspace{-3em}
\end{IEEEbiography}

\begin{IEEEbiography}[{\includegraphics[width=0.8in,height=1in,clip,keepaspectratio]{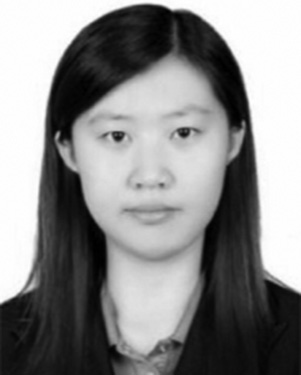}}]{Weili Guan} is now a Ph.D. student with the Faculty of Information Technology, Monash University Clayton Campus, Australia. Her research interests are multimedia computing and information retrieval. 
\end{IEEEbiography}

\begin{IEEEbiography}[{\includegraphics[width=0.8in,height=1in,clip,keepaspectratio]{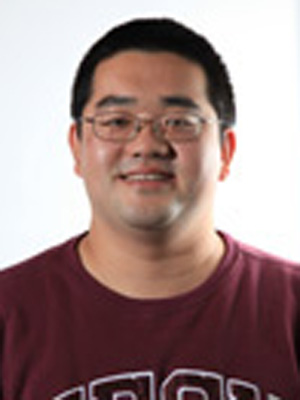}}]{An-An Liu} is a full professor in the school of electronic engineering, Tianjin University, China. He received his Ph.D degree from Tianjin University in 2010. His research interests include computer vision and machine learning.
\end{IEEEbiography}

\begin{IEEEbiography}[{\includegraphics[width=0.8in,clip,keepaspectratio]{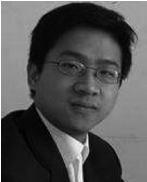}}]{Meng Wang} received the BE and PhD degrees in the Special Class for the Gifted Young and the Department of Electronic Engineering and Information Science from the University of Science and Technology of China (USTC), Hefei, China, respectively. He is a professor at the Hefei University of Technology, China. His current research interests include multimedia content analysis, search, mining, recommendation, and large-scale computing. He received the best paper awards successively from the 17th and 18th ACM International Conference on Multimedia, the best paper award from the 16th International Multimedia Modeling Conference, the best paper award from the 4th International Conference on Internet Multimedia Computing and Service, and the best demo award from the 20th ACM International Conference on Multimedia.
\vspace{-3em}
\end{IEEEbiography} 




\end{document}